\documentclass[journal]{IEEEtran}

\usepackage[utf8]{inputenc} % allow utf-8 input
\usepackage[T1]{fontenc}    % use 8-bit T1 fonts
\usepackage{url}            % simple URL typesetting
\usepackage{booktabs}       % professional-quality tables
\usepackage{amsfonts}       % blackboard math symbols
\usepackage{nicefrac}       % compact symbols for 1/2, etc.
\usepackage{microtype}      % microtypography
\usepackage{lipsum}
\usepackage{times}
\usepackage{epsfig}
\usepackage{graphicx}
\usepackage{epstopdf}
\usepackage{amsmath}
\usepackage{amssymb}
\usepackage{tabularx}
\usepackage{bm}
\usepackage{multirow}
\usepackage{booktabs}
\usepackage[linesnumbered,ruled,vlined ]{algorithm2e}

\usepackage{color}
\usepackage{subfig}
\usepackage{float}

\DeclareMathOperator*{\argmax}{arg\,max}

\newcommand{\tW}{\widetilde{W}}
\newcommand{\m}{\mathrm{max}}
\newcommand{\RID}{\mathrm{RID}}

\begin{document}
\title{Cryo-Electron Microscopy Image Analysis Using Multi-Frequency Vector Diffusion Maps}

\author{Yifeng Fan, Zhizhen~Zhao
\thanks{YF is with the Department
of Electrical and Computer Engineering, University of Illinois at Urbana-Champaign, Urbana, IL, 61820 USA e-mail: yifengf2@illinois.edu }
\thanks{ZZ is with the Department
of Electrical and Computer Engineering, University of Illinois at Urbana-Champaign, Urbana, IL, 61820 USA e-mail: zhizhenz@illinois.edu .}% <-this % stops a space
}

\maketitle

\begin{abstract}
Cryo-electron microscopy (EM) single particle reconstruction is an entirely general technique for 3D structure determination of macromolecular complexes. However, because the images are taken at low electron dose, it is extremely hard to visualize the individual particle with low contrast and high noise level. In this paper, we propose a novel approach called multi-frequency vector diffusion maps (MFVDM) to improve the efficiency and accuracy of cryo-EM 2D image classification and denoising. This framework incorporates different irreducible representations of the estimated alignment between similar images. In addition, we propose a graph filtering scheme to denoise the images using the eigenvalues and eigenvectors of the MFVDM matrices. Through both simulated and publicly available real data, we demonstrate that our proposed method is efficient and robust to noise compared with the state-of-the-art cryo-EM 2D class averaging and image restoration algorithms.
\end{abstract}

\section{Introduction}
\label{sec:intro}
\begin{figure}[htb!]
    \centering
    \includegraphics[width = 1.0\columnwidth]{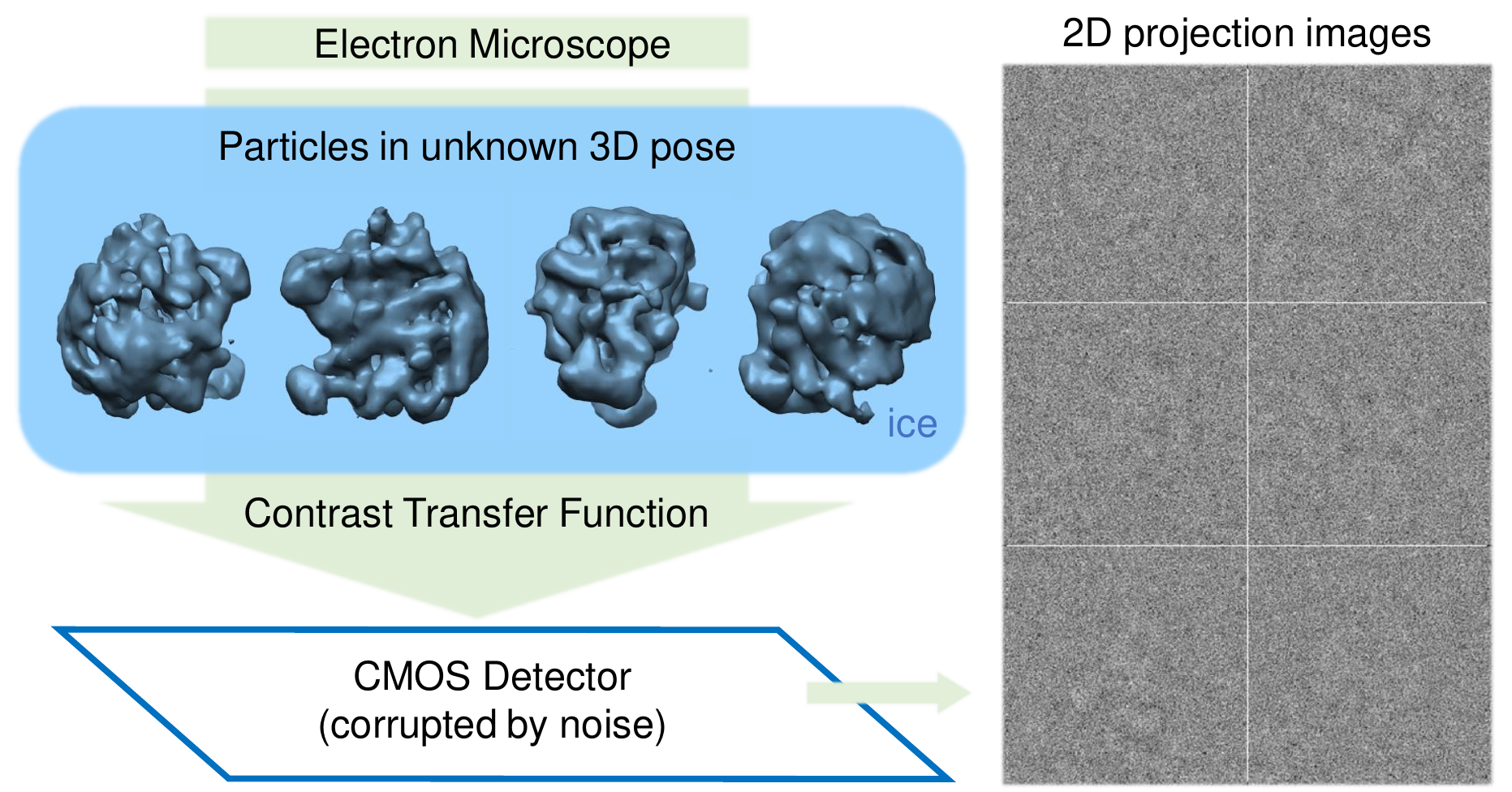}
    \caption{Illustration of image formation in cryo-EM: The electron beam projects the randomly oriented particles. The projection image is modulated by the contrast transfer function (CTF) and corrupted by noise with extremely low SNR.}
    \label{fig:Cryo_EM_illustration}
\end{figure}

%Problem cryo-EM
%A picture is a key to understanding. 
Scientific breakthroughs often build upon successful visualization of objects invisible to the human eye. The 2017 Nobel Prize in Chemistry was awarded for the development of cryo-electron microscopy (cryo-EM) that allows us to image biomolecules in their native functional states at high resolution. This method has moved biochemistry into a new era. 
In cryo-EM experiments, the purified protein particles are distributed evenly within the special sample grid holes assuming a variety of random orientations (see Fig.~\ref{fig:Cryo_EM_illustration}). 
%The grid is flash-frozen at liquid ethane temperature to trap the particles in a thin film of vitreous ice, capturing the native protein structure at the moment of freezing and protecting the sample to some degree from radiation damage. The micrographs that contain the 2D projections of the particles are then collected from the electron microscope. Because of the low electron dose applied in the experiments, the projection images are extremely noisy. Therefore, it requires collecting and processing a large amount of single particle images to achieve high resolution for building the atomic models. 
The single particle images are boxed out during the particle picking step~\cite{langlois2014automated,heimowitz2018apple}. Despite the recent progress, there still remain a lot of challenges for the current computational framework of cryo-EM single particle reconstruction (SPR). For example, in its present form, the methodology still cannot resolve high resolution structures of small particles with molecular weights below 50kDa due to low contrast and signal-to-noise ratio (SNR)~\cite{sigworth2016principles}.

To improve the image quality, a crucial step is the alignment and averaging of the 2D projection images, a procedure known as ``2D class averaging.'' The 2D analysis of cryo-EM data is used to assess the data quality. They can be used for direct observation to look for heterogeneity, discover symmetry, and separate particles into subgroups for additional analysis. Better denoising also increases the diversity of the selected particle views during particle picking, which is crucial for 3D reconstruction. The denoised images are used for common-lines based 3D \textit{ab initio} reconstruction~\cite{wang2013orientation}. A good initial model reduces the number of refinement iterations and leads to a better reconstruction~\cite{zhao2014rotationally}. Alternative 3D \textit{ab initio} modeling method uses higher-order moments of the 2D images, such as sample mean, covariance, and triple correlation~\cite{kam1980reconstruction}. Moment estimation is sensitive to noise, therefore better image denoising is crucial for moment based 3D reconstruction. Therefore, it is important to have fast and accurate algorithms for the 2D analysis and denoising. 

Many existing cryo-EM single particle reconstruction software packages~\cite{scheres2012relion,de2013xmipp,tang2007eman2} contain algorithms for 2D class averaging. %An extensive comparison among different algorithms was reported in~\cite{zhao2014rotationally} and the 
Most common methods use K-means clustering with iterative rotational alignment and clustering. On the other hand, the ASPIRE class averaging pipeline in~\cite{zhao2014rotationally} performs robust rotational invaraint nearest neighbor search and alignment estimation. It uses steerable principal component analysis (sPCA) to efficiently denoise the large image dataset and uses rotationally invariant features to identify images of similar views. For extremely noisy images, the initial nearest neighbor classification has many outliers and it uses vector diffusion maps (VDM)~\cite{singer2011viewing,singer2012vector} to further improve the nearest neighbor classification by taking into account the consistency of in-plane rotations along multiple paths that connect neighboring images through their common neighbors. The improved nearest neighbors are aligned and averaged to form class averaged images that enjoy higher SNR. However, VDM might fail at lower SNRs and generating class averages by rotational and shift alignment is time consuming and prone to interpolation error. %takes into account the both the affinity and the transformation between high-dimensional data points.    %This classification method is called vector diffusion maps (VDM)~\cite{singer2011viewing,singer2012vector}. 
In this paper, we introduce a new algorithm called \emph{multi-frequency vector diffusion maps} (MFVDM) to improve the efficiency and accuracy of cryo-EM 2D image classification and denoising. Based on the initial nearest neighbor identification and alignment, we find including different irreproducible representations of $\mathrm{SO}(2)$ can improve the nearest neighbor classification and alignment estimation. To denoise the images, we use the eigenvalues and eigenvectors of the MFVDM matrices to filter the Fourier-Bessel expansion coefficients of the images and reconstruct the images from the denoised expansion coefficients. To demonstrate our method, we perform comparions with the state-of-the-art algorithms for denoising on two real experimental datasets and simulated datasets at different SNRs. %We detail the results of numerical experiments for simulated projection images of the 70S ribosomal with the purpose of benchmarking the efficiency and accuracy of our new algorithm compared to the state-of-the-art algorithms. The method is also compared with the state of the art on two real datasets. 

%The prior work that we rely on are the truncated fast Fourier-Bessel expansion of the 2D images~\cite{zhao2016fast}, and initial classification in~\cite{zhao2014rotationally} for the similar view identification (determines $w_{ij}$) and the in-plane rotational alignment (determines $\alpha_{ij}$). 
%The novelties of this paper are: 
%(1) Introduce a new metric to identify images of similar views in the presence of high level of noise. For this, we consider path integrals of different irreducible representations of the group transformations. This is in contrast to VDM which uses a single irreducible representation. %, and extend the VDM embedding to $e$VDM  using  eigenvalues and eigenvectors of  $e$VDM matrices. 
% $e$VDM improves the classification and alignment results over VDM significantly. 
% (2) Use the eigenvectors and eigenvalues of $e$VDM matrices to filter the Fourier-Bessel expansion coefficients $a_{k, q}$, which are then used to reconstruct the denoised images. Previously VDM was only used to improve nearest neighbor search and alignment, and the nearest neighbors are registered and averaged to form class averages. The eigenvectors and eigenvalues of the VDM matrix are not used for denoising, since the change of pixel values under image rotation is not a simple phase shift. %, since it is not feasible. 
% It is crucial to use different irreducible representations of $SO(2)$ and filter $a_{k, q}$ instead. 

\section{Related Work}
\label{sec:rel_work}
%Related work, talk about RELION algorithm Maximum likelihood, K-means
In the following we review the cryo-EM 2D class averaging and denoising methods. Traditional 2D class averging uses unsupervised classification methods, such as K-means clustering and reference-free alignment~\cite{frank2006three}.
%2D class averaging and denoising methods are commonly used in cryo-EM image restoration. They start from using unsupervised classification methods, such as K-means clustering~\cite{frank2006three}. 
However, since the images can be in-plane rotated, they cannot be aligned well without being first separated into distinct classes. On the other hand, it is hard to separate classes when the images are not aligned well~\cite{scheres2007modeling}. %Therefore the approaches that iterate between alginment and classification~\cite{frank2006three} aim to address this chicken and egg problem.
%Due to the existance of unknown inplane rotations, the performance of such methods greatly rely on the accuracy of alignments within images~\cite{scheres2007modeling}. 
%However, images cannot be aligned well without a good classification result. 
This leads to a `chicken and agg' problem and approaches that iterate between alginment and classification~\cite{frank2006three} aim to solve it, including maximum-likelihood (ML) estimation~\cite{sigworth1998maximum,scheres2005maximum,scheres2010maximum,lyumkis2013likelihood} and the neural network for kernel probability density estimator self-organizing map (KerDenSOM)~\cite{pascual2001novel}. These methods share a drawback of the high sensitivity to initialization. Therefore, to avoid this, \cite{zhao2014rotationally} introduced rotationally invariant features to identify nearest neighbors and the corresponding alignments. It then uses a nonlinear dimensionality reduction method, VDM~\cite{singer2012vector}, to improve the nearest neighbor search. Images are denoised by a weighted average of their aligned nearest neighbors.

\begin{figure*}[t!]
    \centering
    \includegraphics[width = 1.0\textwidth]{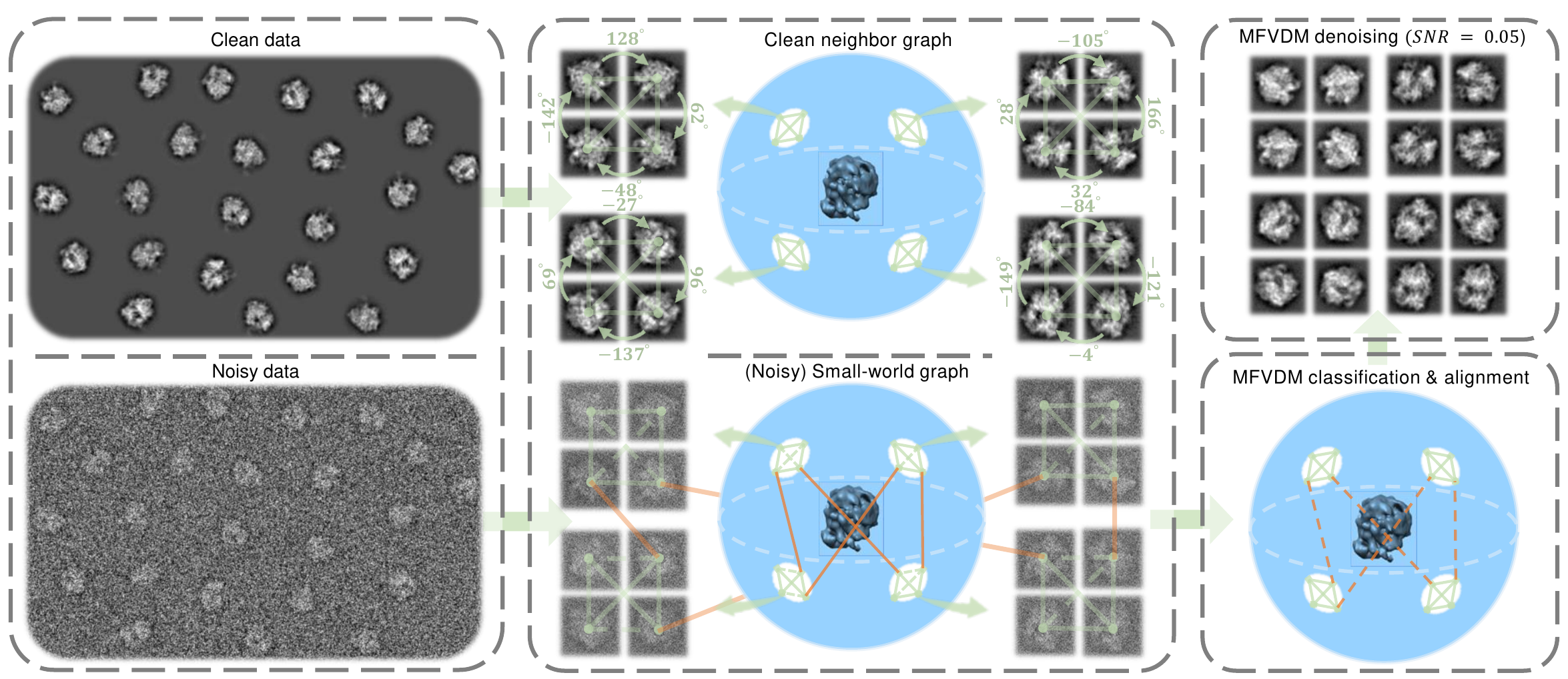}
    \caption{Illustration of MFVDM pipeline: \textit{Left: }Noisy (clean) data samples of a single particle are collected. \textit{Middle:} The distance $d_{\RID}$ defines a neighborhood graph (clean data) and a small world graph (noisy data) on the unit 2-sphere. Among local neighborhood images, the alignments are consistent for clean data.%build the underlying graph with each node as a sample, edge connection is determined by the rotational invariance distance (RID) between image pairs. clean images are locally well aligned and classified, while at noisy case, incorrect classification and alignment happens and leads to shortcuts (orange solid lines) between distant pairs. 
    \textit{Right: }Using multiple frequency representation, MFVDM is able to remove shortcuts (orange dash lines) and improve the alignment accuracy, then graph filters are used for denoising.}
    \label{fig:eVDM_illustration}
\end{figure*}

Another type of approaches aims to denoise EM images without 2D classification and averaging and contains two main methods: (1) steerable PCA (sPCA)~\cite{zhao2013fourier,zhao2016fast} and (2) covariance Wiener filtering (CWF)~\cite{bhamre2016denoising}. SPCA computes the covariance matrices using the truncated expansion coefficients of the noisy images on steerable basis~\cite{zhao2016fast,landa2017steerable} and uses a Wiener-type filtering to denoise the coefficients. %However it does not take into account the effects of contrast transfer by the EM lens. 
For applications to real cryo-EM images, which are convolved by the point spread function of the electron microscope lens, it requires a data preprocessing step ``phase flipping''~\cite{penczek2010image} and does not correct for the phase amplitudes. 
The second method CWF improves the sPCA by estimating the mean and covariance matrix of the clean data without the lens effect, followed by denoising and deconvolution to correct the Fourier phases and amplitudes of the images. However, neither sPCA nor CWF take into account the geometric structure of the projection images as illustrated in Fig.~\ref{fig:eVDM_illustration}. 

Recently, steerable graph Laplacian (SGL)~\cite{landa2018steerable} uses a kernel matrix defined on pairs of images including all rotated versions. The eigenvectors of the kernel matrix correpsond to steerable manifold harmonics on the group invariant data manifold. The denoising step uses the truncated global basis to filter the steerable coefficients of the EM images. By taking into account of the nonlinear geometric structure of the data, it improves the denoising performance compared to sPCA. The assumption that the data lie close to a low dimensional manifold was widely used in natural image analysis and computer vision problems. Various global and local filtering have been proposed~\cite{tomasi1998bilateral,buades2005non,yang2010image,giannakis2012symmetries, schwander2012symmetries,osher2017low,yin2017tale,chan2018performance} for image denoising, inpainting, and super-resolution. 

\begin{figure*}[t!]
	\centering
	\includegraphics[width= 0.8\textwidth]{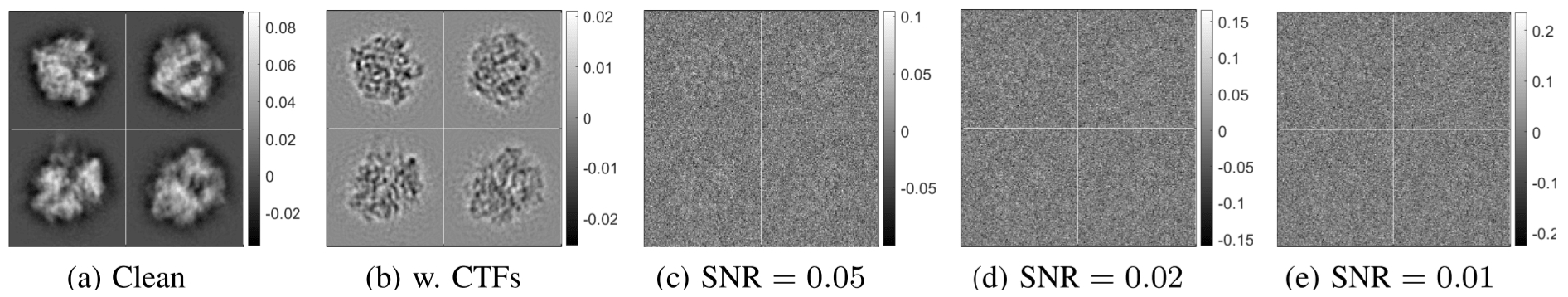}
	\caption{Simulated 70S ribosome projections images: (a) Clean projection images. (b) Clean images altered by CTFs. (c) to (e) Images contaminated by additive white Gaussian noise at $\text{SNR} = 0.05, 0.02, 0.01$.}
	\label{fig:clean_noisy}
\end{figure*}

%\begin{figure*}[t!]
%\centering
%\captionsetup[subfigure]{oneside,margin={-0.5cm,0cm}}
%\subfloat[Clean]{\includegraphics[width = 0.145\textwidth]{figures/sim_data/clean.png}}\;
%\subfloat[w. CTFs]{\includegraphics[width = 0.145\textwidth]{figures/sim_data/clean_CTF_new.png}}\;
%\subfloat[$\text{SNR} = 0.05$]{\includegraphics[width = 0.145\textwidth]{figures/sim_data/noisynoisy_CTF_0_05.png}}\;
%\subfloat[$\text{SNR} = 0.02$]{\includegraphics[width = 0.145\textwidth]{figures/sim_data/noisynoisy_CTF_0_02.png}}\;
%\subfloat[$\text{SNR} = 0.01$]{\includegraphics[width = 0.14\textwidth]{figures/sim_data/noisynoisy_CTF_0_01.png}}
%
%\caption{Simulated 70S ribosome projections images: (a) Clean projection images. (b) Clean images altered by CTFs. (c) to (e) Images contaminated by additive white Gaussian noise at $\text{SNR} = 0.05, 0.02, 0.01$.}
%\label{fig:clean_noisy}
%\end{figure*}

\section{Methods}
\label{sec:methods}

\subsection{Image Formation Model and Geometry}
\label{sec:IFM}
%clean projection image formation
In cryo-EM single particle experiments, each protein sample is embedded in  a thin ice layer at an unknown orientation characterized by a $3 \times 3$ orthogonal matrix  $ R \in \text{SO(3)}$, ($R = [ R^1, R^2, R^3 ]$). %, with respect to the direction of the microscope beam. 
The single particle image $I$ is formed as the X-ray transform of the underlying 3D electron density of the macromolecule. Excluding the contribution of noise, the intensity $I(x, y)$ of the pixel located at $(x, y)$ in the image plane corresponds to the line integral, 
$I(x, y) = \int \limits_{-\infty}^\infty \phi(x R^1 + y R^2 + z R^3) dz$,
where $\phi$ is an unknown 3D volume in a fixed coordinate system. 
%induced by the molecule along the path of the imaging electrons a
%\begin{equation}
%I(x, y) = \int \limits_\infty^\infty V(x R^1 + y R^2 + z R^3) dz
%\end{equation}
For the image $I$, the matrix  $R$ can be divided into two parts: (1) the last column  $ R^3 $ represents the viewing direction $ v =  R^3 $ and uniquely defines a point on a 2-dimensional quotient manifold $\mathcal{M} =  \mathrm{SO}(3)/\mathrm{SO}(2)$ (2-dimensional unit sphere) for a generic volume without symmetry; (2) the first two columns of  $ R $, namely  $ R^1$ and $ R^2 $, form a basis of the tangent plane $T_{I} \mathcal{M}$ and they correspond to the $x, y$ coordinates in the 2D projection image. 

%If the particles are well centered, each projection image $I$ is a noisy view of the macromolecule with an unknown latent $3 \times 3$ orthogonal matrix  $ R \in \text{SO(3)}$, ($R = [ R^1, R^2, R^3 ]$) relative to the direction of the microscope beam. 
 %

% Talk about the noise model
The image is corrupted by two phenomena: a contrast transfer function (CTF)~\cite{frank2006three} and noise. The CTF is analogous to the effects of defocus in a conventional light camera. 
It modulates the phase amplitudes and flips the signs of the phases in the Fourier domain, which leads to a systematic alteration of the image data (see Figs.~\ref{fig:clean_noisy}(a) and \ref{fig:clean_noisy}(b)). The parameters for CTFs are assumed to be known or estimated through standard tools. %and an elementary CTF correction is `phase flipping'}. 
A simple CTF correction method that does not affect the noise statistics is called phase flipping~\cite{penczek2010image}, which corrects the sign of the phases, but not its ampltitude. In our framework, we preprocess the raw images by phase flipping. 

% \frank{The contrast transform function (CTF) $C_i$ that applies on projection image $I_i$ can be modeled on Fourier domain as:
% \begin{equation}
%     I_i(x,y) = C_i(x,y)H_i(x,y)
% \end{equation} where $H_i$ is the true projection on Fourier domain.} 
For noisy particle images (see Fig.~\ref{fig:clean_noisy}(c)), 2D class averaging is an important step to denoise and check the quality of the dataset. It first identifies images with similar views and registers those images by in-plane rotation and shifts. In this paper, we assume that the particles are well centered, and later in the experimental results, we show that the algorithms are robust to small shifts.

For a pair of images  $I_i$ and $I_j $, it is natural to use the rotational invariant distance (RID) $d_{\RID} (i, j)$ to identify images of similar views,
\begin{equation}
    d_{\RID} (i, j) = \min_{\alpha \in [0, 2\pi)}\left\|I_i- \mathcal{R}_{\alpha} I_j \right\|,
    \label{eq:RID}
\end{equation}
where $\mathcal{R}_\alpha$ is an operator that rotates image $I_j$ counter-clockwise by an angle $\alpha$.
The associated optimal in-plane rotation alignment is denoted by $\alpha_{ij}$. The underlying alignment angle for images of similar views is determined by $R_i$ and $R_j$ according to~\cite[Eqs. (4) and (5)]{zhao2014rotationally}. The vertices on the blue spheres in Fig.~\ref{fig:eVDM_illustration} depict the viewing directions. In the noiseless case, small $d_{\RID}$ indicates close viewing directions. Therefore, connecting image pairs with small $d_{\RID}$ form a neighborhood graph on the sphere as shown in the upper middle panel in Fig.~\ref{fig:eVDM_illustration}. However, when the images are extremely noisy, noise aligns well instead of the underlying signals. This results in connecting images of distant views (see the shortcut red edges in the lower middle panel of Fig.~\ref{fig:eVDM_illustration}) and error in the estimation of $\alpha_{ij}$.  

To address these challenges, we first use the filtered sPCA coefficients of the images~\cite{zhao2016fast} to compute $d_{\RID}$ and $\alpha_{ij}$, which can withstand certain level of noise. Furthermore, in the following subsections, we introduce a graph based method to robustly learn the nonlinear geometrical structure of the image data. The new pipeline combines multiple irreducible representations of the compact
alignment group with a random walk executed on
a graph where nodes correspond to images and edges denote connections. We then use the learned basis of the manifold to improve nearest
neighbor search, alignment, denoising, and deconvolution.

\subsection{Multi-Frequency Vector Diffusion Maps}
\label{subsec:eVDM}
For clean projection images, since the viewing directions of the images lie on a 2-sphere, by using $d_\RID$, we are able to build a local neighborhood graph $(V,E)$ over the sphere with the nodes $V$ representing the viewing directions and the edges $E$ connecting images of similar views, i.e., $(i, j) \in E$ indicates $\langle v_i, v_j \rangle \approx 1$. All the views of similar images are in a small spherical cap (see the upper middel panel of Fig.~\ref{fig:eVDM_illustration}). %Therefore, the clean graph is a neighborhood graph on the sphere and 
In addition, the rotational alignment has cycle consistency among the neighbors, i.e., summing over angles along the cycles will be close to $0\mod 2\pi$.
 
\begin{figure}[t!]
    \centering
	\subfloat[Affinity]{\includegraphics[width= 0.27\textwidth]{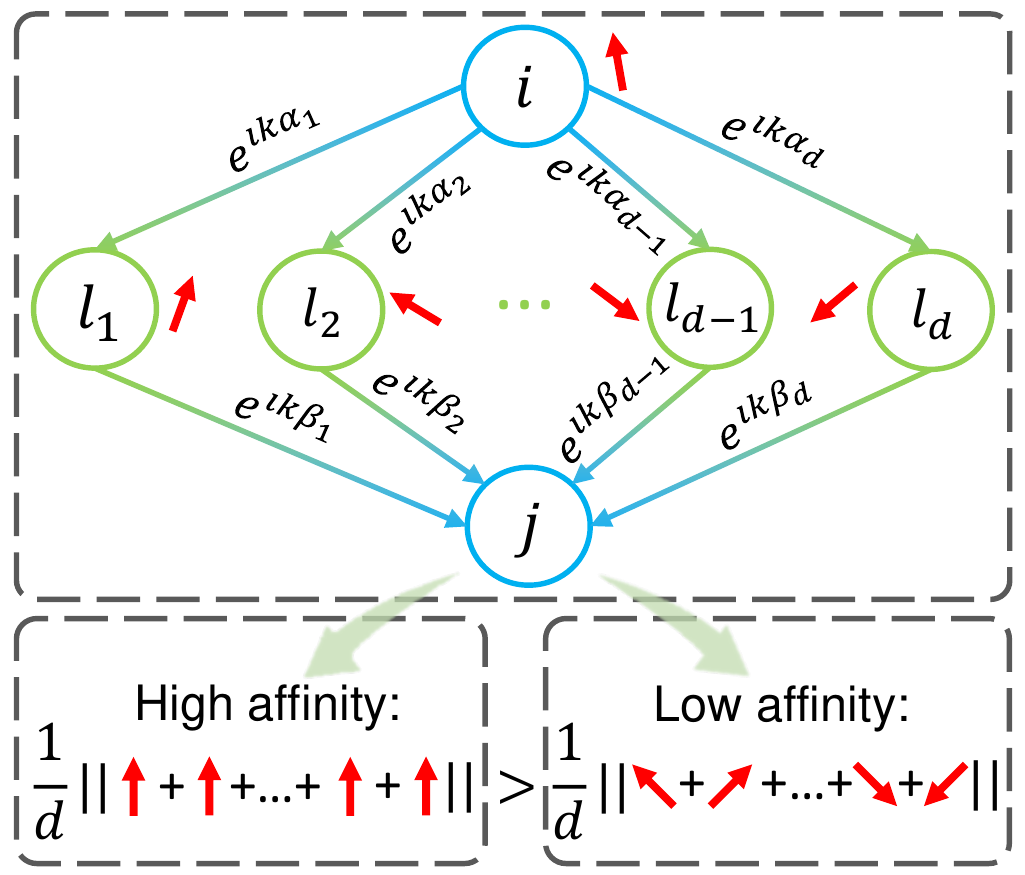}
	\label{fig:eVDM_affinity}
	}
	\subfloat[Alignment]{\includegraphics[width= 0.23\textwidth]{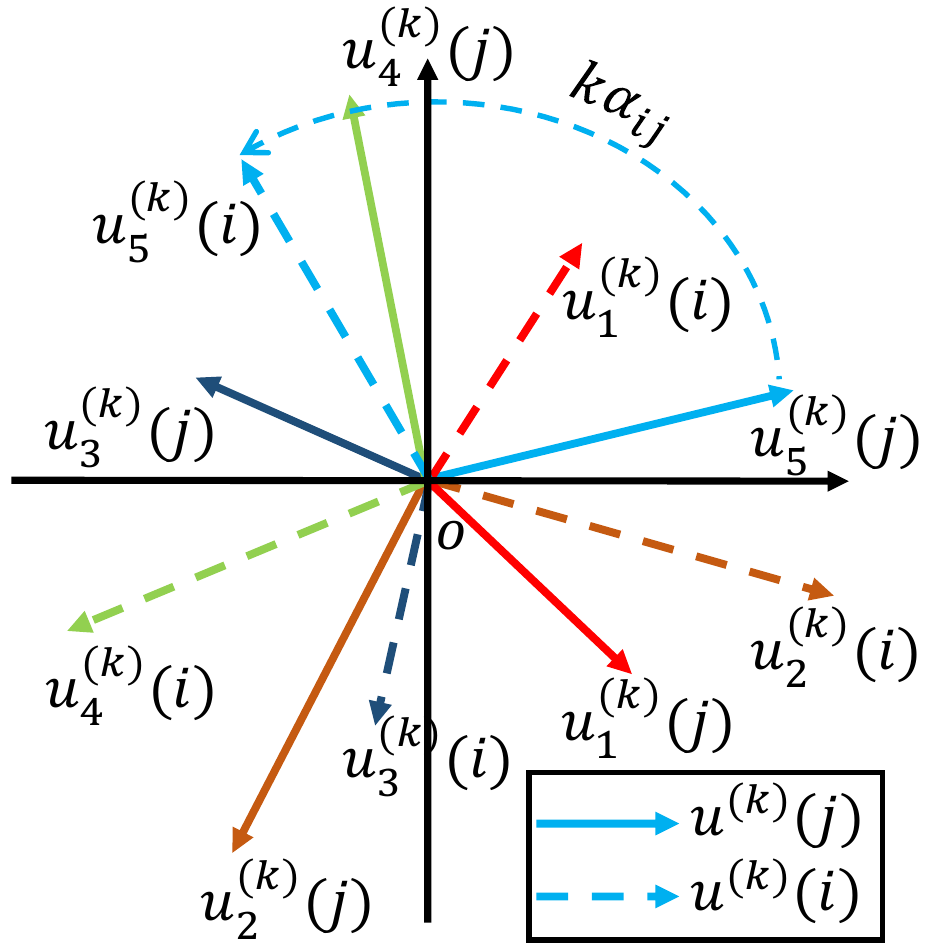}
	\label{fig:eVDM_align}
	}
    \caption{Illustration of MFVDM affinity and alignment. \protect \subref{fig:eVDM_affinity} At each frequency $k$ we average the vectors transported along paths of length 2 connecting $i$ and $j$. If the transportations are consistant, the nodes have high affinity. \protect \subref{fig:eVDM_align} %Rotational alignment illustration. 
    When $v_i = v_j$, the tangent planes $T_{I_i} \mathcal{M}$ and $T_{I_j} \mathcal{M}$ coincide and the eigenvectors of $\tW_k$ satisfy Eq.~\eqref{eq:sameview}.}
    \label{fig:eVDMalign}
\end{figure}

% naturally we build a graph $(V,E)$ upon the sPCA initial classification and in-plane rotational alignment described above. Each node represents an image and the edges connect similar pairs of images. i.e., $(i, j) \in E$ if $i$ is $j$'s nearest neighbor or $j$ is $i$'s nearest neighbor. 
% With initial rotational invariant nearest neighbors and the corresponding in-plane rotational alignment angles, we can build a graph with each image as a node in the graph and the edges connect nearest neighbor pairs, i.e., $(i, j) \in E$ if $i$ is $j$'s nearest neighbor or $j$ is $i$'s nearest neighbor. 
Based on this graph structure for $n$ images, we build a class of $n \times n$ Hermitian matrices $W_k$, for $k = 0,1, \dots, \tilde{k}$,
\begin{equation}
W_k(i, j) = \begin{cases} 
e^{\imath k \alpha_{ij}} & (i, j) \in E, \\ 
0 & (i, j) \notin E,
\end{cases}
\label{eq:Wk}
\end{equation}
where $\alpha_{ij} = -\alpha_{ji}$. The degree of each node $\deg(i) = \sum_{j: (i, j) \in E } 1$. We define a diagonal matrix $D$ of the same size as $W_k$ with $D(i, i) = \deg(i)$. Then the matrix $S_k = D^{-1} W_k$ is applied to vectors $u$ in the complex plane, which can be viewed as an averaging operator for the tangent vector fields, since
\begin{equation}
(S_k u)(i) = \frac{1}{\deg(i)} \sum_{j: (i, j) \in E} e^{\imath k \alpha_{ij}} u(j). 
\label{eq:av}
\end{equation}
The operator $S_k$ transports vectors from the tangent spaces $T_{I_j} \mathcal{M}$ to $T_{I_i} \mathcal{M}$, for all nearest neighbor $j$, and then averages the transported vectors in $T_{I_i} \mathcal{M}$ (see Fig.~\ref{fig:eVDM_affinity}). %\textcolor{red}{It plays a similar role as the class average process.}

%In the eVDM framework, the affinity between images $i$ and $j$ are measured by considering all paths of length $t$ connecting them. We sum over the edge weights and the transformations under different irreducible representations of $\mathrm{SO}(2)$ (see Figs.~\ref{fig:eVDMaffhigh} and~\ref{fig:eVDMafflow}). Every path from $i$ to $j$ may result in a different transformation, for example, parallel transport due to curvature of the manifold or wrong identification of nearest neighbors due to noise. When adding the transformations under a certain irreducible representation along different paths, cancellations will happen. 

\begin{algorithm}[t!]
\SetAlgoLined
\KwIn{Nearest neighbors list and alignment angles for noisy cryo-EM images from initial classification in~\cite{zhao2014rotationally}}
\KwOut{$s$-nearest neighbors for each image and the corresponding optimal alignments $\hat{\alpha}_{ij}$, for $(i, j)\in E$}
\For{$ k = 1, \dots, \tilde{k} $}{
Construct the normalized affinity matrix $\tW_k$ according to Eqs.~\eqref{eq:Wk} and~\eqref{eq:tW}\\
Compute the largest $m$ eigenvalues $\lambda^{(k)}_1 \geq \lambda^{(k)}_2, \geq, \dots, \geq \lambda^{(k)}_m$ of $\tW_k$ and the corresponding eigenvectors $\{ u^{(k)}_l \}_{l = 1}^m$ \\
Compute the truncated MFVDM embedding $V^{(k)}_{t, m}$ according to Eq.~\eqref{eq:eVDMmapT} \\
}
Compute the normalized affinity using Eq.~\eqref{eq:EVDMaff} \\
Identify $s$ nearest neighbors for each data point \\
Compute $\hat{\alpha}_{ij}$ for nearest neighbor pairs using Eq.~\eqref{eq:EVDMalign2}
\caption{MFVDM nearest neighbor search and alignment}
\label{alg:EVDMclass}
\end{algorithm}
When the images are noisy, the evaluation of $d_\RID$ is not accurate and creates shortcut edges, resulting in a small world graph~\cite{watts1998collective} (see lower middle panel of Fig.~\ref{fig:eVDM_illustration}) and leads to poor class average process. To address this problem, note the rotational alignment consistency breaks along the shortcut edges, i.e., if $(i, j) \in E$ is a shortcut, the transportations on different paths between $(i, j)$ are inconsistent (see Fig.~\ref{fig:eVDM_affinity}). Therefore, we define the affinity between $(i, j)$ by taking into account the consistency of transporting the vectors rotated at different angular frequencies $k$ along the paths with length $2t$, as $\sum_{k = 1}^{\tilde{k}} |S^{2t}_k(i, j)|^2 $. Since $S_k$ is similar to the symmetric matrix $\tW_k$, 
\begin{equation}
\widetilde{W}_k = D_0^{-1/2} W_k D_0^{-1/2},
\label{eq:tW}
\end{equation}
we can quantify the consistency %in transporting the vector along paths of length $2t$ 
by computing  $\sum_{k = 1}^{\tilde{k}} |\tW_k^{2t}(i, j)|^2$.
%\begin{equation}
%\sum_{k = 1}^{\tilde{k}} \left|S_k^{2t}(i, j) \right|^2 = \frac{\deg(j)}{\deg(i)} %\sum_{k = 1}^{\tilde{k}} \left|\tW_k^{2t}(i, j) \right|^2.
%\label{eq:kVDMaf}
%\end{equation}
Since $\tW_k$ is Hermitian, it has real eigenvalues $\{\lambda^{(k)}_l\}_{l = 1}^n$ in descending order %\frank{Need to mention the order?}
and the associated eigenvectors $\{u^{(k)}_l\}_{l = 1}^n $. %it has the following spectral decomposition:
%\begin{equation}
%\tW_k(i, j) = \sum_{l = 1}^n \lambda^{(k)}_l u^{(k)}_l(i) \overline{u^{(k)}_l(j)}, \quad \text{and} \quad 
%\tW^{2t}_k (i, j) = \sum_{l = 1}^n \lambda_l^{2t} u^{(k)}_l(i) \overline{u^{(k)}_l(j)}.
%\end{equation}
%where $u^{(k)}_l(i) \in \mathbb{C}$, for $i = 1, \dots, n$ and $l = 1, \dots, n$, and  %.$ \left| \tW_k^{(2t)}(i, j) \right| $ is calculated using the eigenvalues and eigenvectors: 
%\begin{equation}
%\left|\tW_k^{2t}(i, j) \right|^2 = \sum_{l, r = 1}^n (\lambda^{(k)}_l \lambda^{(k)}_r)^{2t} \langle u^{(k)}_l(i), u^{(k)}_r(i) \rangle \langle u^{(k)}_l(j), u^{(k)}_r(j) \rangle. 
%\label{eq:trHS}
%\end{equation}
%Therefore, the affinity is an inner product for the finite dimensional Hilbert space $\mathbb{C}^{n^2}$ via the mapping: 
%\begin{equation}
%V_t : i \mapsto \left( (\lambda_l, \lambda_r)^t\langle v_l(i), v_r(i) \rangle \right)_{l, r = 1}^n
%\end{equation}
%The vector diffusion distance between nodes $i$ and $j$ is denoted $d_{\VDM, t}(i, j)$ and is defined as 
%\begin{equation}
%d_{\VDM, t}^2(i, j) = \| V_t(i) - V_t(j) \|_2^2.
%\label{eq:distVDM}
%\end{equation}
%It was shown in~\cite{singer2012vector} that the vector diffusion mapping and distances can be well approximated by using only the few largest eigenvalues and their corresponding eigenvectors and was applied to cryo-EM 2D class averaging~\cite{zhao2014rotationally} to improve the nearest neighbor identification.
%\frank{In this way, the eVDM mapping is defined as} 
With the MFVDM mapping defined as 
\begin{equation}
V^{(k)}_t : i \mapsto \left( (\lambda^{(k)}_l \lambda^{(k)}_r)^t\langle u^{(k)}_l(i), u^{(k)}_r(i) \rangle \right)_{l, r = 1}^n, 
\label{eq:kVDMmap}
\end{equation}
we have $\left|\tW_k^{2t}(i, j) \right|^2 = \left \langle V^{(k)}_t(i), V^{(k)}_t(j) \right \rangle $ for $k = 1, \dots, \tilde{k}$. Through the nonlinear-embedding in~\eqref{eq:kVDMmap}, the similarity of two images can be evaluated by the inner product of the embedded vectors. 
%The $e$VDM distance is defined as, 
%\begin{equation}
%d_{\eVDM, t}^2(i, j) = \sum_{k = 1}^{\tilde{k}} \left \| V^{(k)}_t(i) - V^{(k)}_t(j) \right \|_2^2.
%\label{eq:distEVDM}
%\end{equation}
VDM~\cite{singer2012vector} is a special case of MFVDM with $\tilde{k} = 1$. It was shown in~\cite{singer2012vector} that the VDM can be well approximated by using only the few largest eigenvalues and their corresponding eigenvectors. Similarly, %$e$VDM can use only the top $m$ eigenvectors and eigenvalues for each $k$. Therefore, 
we truncate MFVDM mapping as
\begin{equation}
V^{(k)}_{t, m} : i \mapsto \left( (\lambda^{(k)}_l \lambda^{(k)}_r)^t\langle u^{(k)}_l(i), u^{(k)}_r(i) \rangle \right)_{l, r = 1}^m,  
\label{eq:eVDMmapT}
\end{equation}
with $m < n$. When $t$ is small and $m$ and $\tilde{k}$ are chosen properly~\cite{lin2016embeddings}, the embedding is able to preserve the topological structure of the underlying data manifold and is robust to noise perturbation. 

\vspace{0.1cm}
\noindent\textbf{Nearest Neighbor Search: }
Based on the truncated mapping, we identify nearest neighbors by the normalized affinity $A$ defined as
\begin{equation}
A(i, j) = \sum_{k = 1}^{\tilde{k}} \left \langle \frac{V^{(k)}_{t, m}(i)}{\|V_{t,m}^{(k)}(i)\|}, \frac{V^{(k)}_{t,m}(j)}{\|V_{t,m}^{(k)}(j)\|} \right \rangle.
\label{eq:EVDMaff}
\end{equation}
%, which is empirically more robust to noise than the $e$VDM distance in Eq.~\eqref{eq:distEVDM}. 
Since we take into account various irreducible representations of $\mathrm{SO}(2)$ and the consistency of the corresponding transformations over the graph, this new affinity in Eq.~\eqref{eq:EVDMaff} is able to avoid shortcuts, and is more robust to noise compared to VDM (see Fig.~\ref{fig:classification_rot}).
%The $s$ nearest neighbors are identified using the normalized eVDM affinity in~\eqref{eq:EVDMaff}. 

\vspace{0.1cm}
\noindent\textbf{Rotational Alignment: }To estimate the in-plane rotational alignment angles for images of similar views, we notice that the eigenvectors of $\tW_k$ encode the alignment information between neighboring images, as illustrated in Fig.~\ref{fig:eVDM_align}. Specifically, for images of the same views $v_i = v_j$, %we have $u^{(k)}_l(i) = e^{\imath k \alpha_{ij}} u_l^{(k)}(j)$, for $l = 1, \dots, n$. 
the corresponding entries of eigenvalues are vectors in the complex plane and, 
\begin{equation}
u^{(k)}_l(i) = e^{\imath k \alpha_{ij}} u_l^{(k)}(j), \quad \forall\, l = 1, \dots, n. 
\label{eq:sameview}
\end{equation}
When the viewing directions are close but not identical, Eq.~\eqref{eq:sameview} holds approximately. 
%The entries $u_l^{(k)}(i)$ deviates from $e^{\imath k \alpha_{ij}} u_l^{(k)}(j)$ as the eigenvalues $\lambda^{(k)}_l$ become smaller, since the eigenvectors become more oscillatory and more sensitive to noise. 
Therefore, we minimize the mean squared error between $\lambda_l^{(k)} u_l^{(k)}(i)$ and $e^{\imath k\alpha}\lambda_l^{(k)} u_l^{(k)}(j)$ for all $l$ and $k$ to estimate the in-plane rotational alignment, %which is equivalent to maximizing the weighted cross-correlation,
%\begin{equation}
%
, which can be simplified to 
%\label{eq:EVDMalign1}
%\end{equation}
%We rewrite Eq.~\eqref{eq:EVDMalign1} into 
\begin{equation}
\hat{\alpha}_{ij} = \argmax_\alpha \sum_{k = 1}^{ \tilde{k} } \sum_{l = 1}^m \left(\lambda_l^{(k)}\right)^{2t} u_l^{(k)}(i) \overline{u_l^{(k)}(j)} e^{-\imath k \alpha}. 
\label{eq:EVDMalign2}
\end{equation}
The optimal alignment can be approximated by locating the peak of the FFT of the zero-padded $z$, where $z(k) = \sum_{l = 1}^m \left(\lambda_l^{(k)}\right)^{2t} u_l^{(k)}(i) \overline{u_l^{(k)}(j)}$. 
%Let $n_{ij} = \argmax{n} \sum_{-N}^{N}u(k) e^{-\imath 2\pi \frac{n k}{N}}$ and $\alpha^*_{ij} = 2\pi \frac{n_{ij}}{N}$. %Alternatively, we define $\tilde{u}_k = \frac{u_k}{|u_k|}$ and use FFT of zero padded $\tilde{u}$. \textcolor{red}{Need to check which one gives better alignment, with or without normalization.}  
%\vspace{-1cm}
%\begin{figure*}[t!]
%\begin{center}
%\subfloat[high affinity]{	
%	\includegraphics[width = 0.32\textwidth]{./figures/eVDM_1}
%	\label{fig:eVDMaffhigh}
%}
%\subfloat[low affinity]{	
%	\includegraphics[width = 0.32\textwidth]{./figures/eVDM_2}
%	\label{fig:eVDMafflow}
%}
%\subfloat[eVDM align]{	
%	\includegraphics[width= 0.25\textwidth]{./figures/rotation_2}
%	\label{fig:eVDMalign}
%}
%\end{center}
%\caption{Illustration for eVDM classification and alignment. \protect \subref{fig:eVDMaffhigh} and \protect \subref{fig:eVDMafflow} use the consistency of transporting vectors along paths of length 2 to define affinity between nodes $i$ and $j$. When vectors from different paths are added together the amplitude of the resulting vector can be as large as the number of paths if they are all consistent \protect \subref{fig:eVDMaffhigh}, or much smaller due to inconsistencies \protect \subref{fig:eVDMafflow}. \protect\subref{fig:eVDMalign} When $v_i = v_j$, the tangent plane at point $i$ coincides with the tangent plane at point $j$ and the eigenvectors of $\tW_k$ satisfy Eq.~\eqref{eq:sameview}.}
%\label{fig:eVDM}
%\vspace{-0.5cm}
%\end{figure*}

The MFVDM nearest neighbor search and alignment algorithm in Alg.~\ref{alg:EVDMclass} is very efficient in terms of runtime and memory requirements, because it is based on the computation of the top eigenvectors of several sparse Hermitian matrices, which can be easily parallelized. 

\subsection{Denoising and CTF Correction}
\begin{algorithm}
\SetAlgoLined
\KwIn{$s$-nearest neighbor list and the corresponding optimal alignment angles $\hat{\alpha}_{ij}$ estimated from Alg.~\ref{alg:EVDMclass}, image and absolute CTF expansion coefficients $a^i_{k, q}$ and $c^i_{k, q}$, for $i = 1, \dots, n$}
\KwOut{Denoised images $\hat{I}_1, \dots \hat{I}_n$}
\For{$k = 0, 1, \dots, k_\m$}{
Construct the normalized affinity matrix $\tW_k$ according to Eqs.~\eqref{eq:Wk} and~\eqref{eq:tW}\\
%Apply the average operator $\left(2 A_k -  A^2_k \right)$ on its corresponding coefficients $\hat{a}_{k,q}$.\\
Compute the top $m$ eigenvectors $\{u^{(k)}_l\}_{l = 1}^m $ and the corresponding eigenvalues $\lambda^{(k)}_1 \geq \lambda^{(k)}_2, \geq, \dots, \geq \lambda^{(k)}_m $ \\
Use the $u^{(k)}_l$ and $\lambda^{(k)}_l$ to filter the image expansion coefficients using Eq.~\eqref{eq:fA} and CTF expansion coefficients.
}
Reconstruct $\hat{I}_1, \dots \hat{I}_n$ using Eqs.~\eqref{eq:denoiseI} and ~\eqref{eq:CTFcorrect}. \\
\caption{MFVDM denoising and CTF correction}
\label{alg:EVDMdenoise}
\end{algorithm}
The digital image $I_i$ can be modeled as discrete samples of an underlying continuous function with bandlimit $\kappa$ and the particle is well concentrated within radius of $R$. Therefore, the Fourier coefficients of the image can be well approximated by the truncated expansion on Fourier-Bessel basis $\psi_\kappa^{k, q}(\xi, \theta)$~\cite{zhao2016fast} on the disk of radius $\kappa$, 
\begin{equation}
\mathcal{F}(I_i)(\xi, \theta) =  \sum_{k = -k_\mathrm{max}}^{k_\mathrm{max}} \sum_{q = 1}^{p_k} a^i_{k, q} \psi_{\kappa}^{k, q}(\xi, \theta). 
\end{equation}
When $I_i$ is rotated by an angle $\alpha$, the expansion coefficients $a^i_{k, q}$ will be transformed to $a_{k, q} e^{-\imath k \alpha}$. Similarly, %\frank{Need a more clear illustration on CTF? Maybe refer a link to an illustration of CTF and the image model. like $y = Ax$ on Fourier domain. } \jane{no space, maybe can be added to supplementary material.} 
for each projection image's absolute CTF $C_i$, the expansion coefficients are $c^{i}_{k, q}$. 

We first denoise the expansion coefficients $a^i_{k, q}$ using the MFVDM eigenvectors and eigenvalues, and reconstruct the denoised images. %Unlike other 2D class averaging approaches, where one need to rotationally align the images of similar viewing directions,
%\begin{equation}
%A^{(k)} = A^{(k)} e^{-\imath k \alpha}, \quad \text{for } k = -k_\m, \dots, k_\m,
%\label{eq:Ak_rot}
%\end{equation}
%and under counter-clockwise rotation and reflection with respect to the $y$-axis, 
%\begin{equation}
%A^{(k)} = \overline{A^{(k)}} e^{-\imath k \alpha}, \quad \text{for } k = -k_\m, \dots, k_\m.
%\end{equation}
%We consider the influence of noise on our datasets from three aspects, first one is the noise of Fourier-Bessel coefficients, another is the wrong nearest neighbors classification $N_i$ that induced by noise, and the last one is the incorrect optimal rotation alignment. In our experiment later we will show the deviation of rotation alignment between calculation and theory can be negligible. Therefore, we concentrate on how to reduce the influence of noise on nearest neighbors classification $N_i$.
%In order to understand this, we treat our average operator $ A_k$ as a random graph matrix, where each pair of nearest neighbors is true with probability $p$, and with probability $1-p$ it is a shortcut between any two vertices. Obviously, the higher probability $p$ means the higher SNR of our images. In extreme case when $p = 0$, each edge of nearest neighbors is randomly connected between any vertices. 
Let us denote by $A^{(k)}$ and $C^{(k)}$ the matrices of the coefficients with angular frequency $k$, obtained by putting $a^i_{k, q}$ and $c^{i}_{k, q}$ for all $i$ and all $q$ into a matrix of size $n \times p_k$, where the rows are indexed by the image number $i$ and the columns are indexed by the index $q$. Applying $S_k$ to $A^{(k)}$ denoises the expansion coefficients by directly averaging the coefficients among its nearest neighbors. However this process over-smoothes the underlying signal. Following~\cite{singer2009diffusion,gilboa2002forward}, we use $2S_k - S_k^2$ as another denoising operator. The operator is related to forward and backward diffusion on the graph that first smoothes and then sharpens the signal defined on the graph. %To understand this process, it is applying the averaging operator to the signal and applying this smoothing operator again to the residual between the observed signal and the averaged one. 
This filter reduces the blurring caused by averaging using $S_k$. %Using 
It is equivalent to applying spectral filters to the eigenvalues, with $h(\lambda) = \lambda$ and $h(\lambda) = 2 \lambda - \lambda^2$ for $S_k$ and $2S_k - S_k^2$ respectively.  
%Let us introduce two filtering functions over the eigenvalues,
%\begin{equation}
%   h_1(\lambda_l) = \lambda_l \quad \text{and} \quad   h_2(\lambda_l) = 2\lambda_l - \lambda_l^2, \, \text{for } l = 1, \dots, n.
%\end{equation}
%The spectral filters correspond to the denoising operator $S_k$ and $2S_k - S_k^2$,  
%In this way, the aforementioned filters
%\begin{align}
%    &S_k  = D_0^{-1/2}\tW  D_0^{1/2} = D_0^{-1/2}\left[ \sum_{l = 1}^{n}  h_1\left(\lambda_l^{(k)}\right)\left( u_l^{(k)}\right)\left( u_l^{(k)}\right)^*\right] D_0^{1/2},  \\
%   & 2 S_k -  S_k^2  = D_0^{-1/2}\left[ \sum_{l = 1}^{n}  h_2\left(\lambda_l^{(k)}\right)\left( u_l^{(k)}\right)\left( u_l^{(k)}\right)^*\right] D_0^{1/2}.
%\end{align}
%The eigenvectors with small eigenvalues are more sensitive to noise, therefore, we can also truncate the eigenvalues by 
%\begin{equation}
%   h_3(\lambda_l) = \begin{cases} \lambda_l & l \leq m \\
%   0 & l>m
%   \end{cases}  \quad \text{and} \quad h_4(\lambda_l) = \begin{cases}
%   2\lambda_l - \lambda_l^2, & l \leq m, \\
%   0 & l>m.
%   \end{cases}
%\end{equation}
%In~\cite{singer2009diffusion} it was shown that operator $2 W_k -  W_k^2$ uses the idea of diffusion to smooth reduce noise and then perform inverse diffusion then applying $A$. 

%Although we can directly apply those filters to denoise the coefficients, the denoising effects are still sub-optimal (see Fig.~\ref{fig:eVDM_filter}). 
Since eigenvectors with smaller eigenvalues are more oscillatory and easily perturbed by noise, it is necessary to truncate with the top $m$ eigenvalues. The filtered expansion coefficients are,
%\begin{equation}
%\hat{a}^{i}_{k, q} = \frac{1}{\sqrt{\deg(i)}} \sum_{j = 1}^n \sum_{l = 1}^{m} h\left(\lambda_l^{(k)} \right) u_l^{(k)}(i) \overline{ u_l^{(k)}(j) } \sqrt{\deg(j)} a^{j}_{k, q}, 
%\label{eq:fA}
%\end{equation}
\begin{equation}
    \hat{A}^{(k)} = D^{-1/2} U_k h(\Lambda_k) U_k^* D^{1/2} A^{(k)},
    \label{eq:fA}
\end{equation}
where $U_k^*$ denotes the complex conjuate transpose of $U_k$. 
%for 
The filtering and truncation of the eigenvalues help denoise the expansion coefficients. Similarly, we can compute the effective CTFs if the same filtering operation are applied to the $C_i$ and get filtered effective CTF coefficients $\hat{C}^{(k)}$ according to Eq.~\eqref{eq:fA}. 

%The denoised image sampled on the Cartesian grid $(x, y)$ in real domain can be efficiently evaluated from the denoised coefficients $\hat{a}^i_{k, q}$, 
%We can use the expanionsion coefficients $c^{i}_{k, q}$ to perform CTF correction. 
%\begin{equation}
%    B^i_{k, q, q'} = c^i_{k, q} \int_{0}^\kappa \psi_\kappa^{k, q}(\xi, 0) \psi_\kappa^{k, q'}(\xi, 0) d \xi 
%\label{eq:B}
%\end{equation}
%This can be evaluated through numerical integration. 

%\begin{equation}
%    \tilde{a}^i_{k, q} = \sum_{k', q'} B^{i}_{k - k', q, q'} \hat{a}^{i}_{k', q'}
%\label{eq:CTF_A}
%\end{equation}

The Fourier coefficients of the denoised image without CTF correction is %\frank{use recover or synthesize? }
recovered from the denoised expansion coefficients,
%\begin{align}
%&\widetilde{I}_i(x, y)  = \sum_{k = -k_\m}^{k_\m} \sum_{q = 1}^{p_k} \hat{a}^i_{k, q} \mathcal{F}^{-1}\left( \psi_\kappa^{k, q} \right)(x, y),  %\\
%& = \sum_{q = 1}^{p_0} \hat{a}^i_{0, q} g_{\kappa}^{0, q}(r_{x, y}) + 2 \mathrm{Re}\left[ \sum_{k = 1}^{k_\m} \sum_{q = 1}^{p_k} \hat{a}^i_{k,q} g_{\kappa}^{k, q}(r_{x, y}) e^{\imath k \theta_{x, y}}\right], \nonumber
%\label{eq:denoiseI}
%\end{align}
\begin{align}
&\mathcal{F}\left(\widetilde{I}_i\right)(\xi_1, \xi_2)  = \sum_{k = -k_\m}^{k_\m} \sum_{q = 1}^{p_k} \hat{a}^i_{k, q}  \psi_\kappa^{k, q}(\xi_1, \xi_2), 
\label{eq:denoiseI}
\end{align}
%where $r_{x, y} = \sqrt{x^2 + y^2}$ and $\theta_{x, y} = \tan^{-1}\left( \frac{y}{x}\right)$. 
where $(\xi_1, \xi_2)$ are located on the Cartesian grid points. For CTF $\widetilde{C}_i(\xi_1, \xi_2) = \sum_{k = -k_\m}^{k_\m} \sum_{q = 1}^{p_k} \hat{c}^i_{k, q}  \psi_\kappa^{k, q}(\xi_1, \xi_2)$.
%The inverse Fourier transform of the Fourier-Bessel basis has analytical expression, therefore, it is easy to evaluate Eq.~\eqref{eq:denoiseI}. 
The MFVDM denoising steps are outlined in Alg.~\ref{alg:EVDMdenoise}. 
%\subsection{Contrast Transfer Function Correction for Cryo-EM}
%\label{sec:CTF}
%\frank{I am struggling on how to introduce CTF that make people easily understand.}
%For a given 

%We use the following method to correct CTF effects on the denoised images form eVDM. Suppose the denoised image is $I$ and $\mathcal{F}(I)(\xi_1, \xi_2)$ denotes the Fourier coefficients at frequency $(\xi_1, \xi_2)$. We can estimate the effective CTF function for image $I_i$, $\widetilde{C}_i(\xi_1, \xi_2)$. 
Based on the effective CTF $\widetilde{C}_i$, we perform the following CTF correction to estimate the underlying projection image,
\begin{equation}
    \hat{I}_i (x, y) = \mathcal{F}^{-1} \left( \frac{\mathcal{F}\left(\widetilde{I}_i \right )(\xi_1, \xi_2)}{\widetilde{C}_i(\xi_1, \xi_2)} \right) (x, y).
    \label{eq:CTFcorrect}
\end{equation}
The main advantage of filtering the expansion coefficients compared to the 2D class averaging in many existing software packages~\cite{de2013xmipp,scheres2012relion} is that it does not explicitly align images of similar views which is computationally expensive and prone to interpolation errors.  

%\begin{equation}
%    \widetilde{\mathrm{CTF}}(\xi_1, \xi_2) = \frac{1}{N}\sum_{i = 1}^N \left(\left|\mathrm{CTF}_i(\xi_1, \xi_2)\right |\right).
%    \label{eq:eCTF}
%\end{equation}
%\begin{align}
%    \widetilde{\mathrm{CTF}}_i(\xi_1, \xi_2) &= \frac{1}{\sqrt{\deg(i)}} \sum_{j = 1}^n \sum_{l = 1}^{m} h\left(\lambda_l^{(k)} \right) u_l^{(0)}(i) \overline{ u_l^{(0)}(j) } \nonumber \\
%    & \quad \quad \times \sqrt{\deg(j)} \left|\mathrm{CTF}_j(\xi_1, \xi_2) \right|,
%    \label{eq:eCTF}
%\end{align}
%The effective CTFs can be evaluated as applying the same global filtering to the CTFs for each individual projections. Also we assume that the CTFs are circularly symmetric\footnote{This can be extended to non-circularly symmetric CTFs}. If we arrange the phase-flipped CTFs $C_i$ for each projection image into a matrix of size $n \times L^2$, with the row number indicating the image number and each row contains a vectorized CTF, then the effective CTF after eVDM denoising is evaluated according to,
%\begin{align}
%    \widetilde{C} = D^{-1/2} U_0 h(\Lambda_0) U_0 D^{1/2} \mathrm{C}
%    \label{eq:eCTF}
%\end{align}
%The $i$th row of the matrix $\widetilde{C}$, denoted as $\widetilde{C}_i$ is the effective CTF for image $I_i$. 

%We assume that the denoised images from eVDM correspond to the original projection images modified by the effective CTF in~\eqref{eq:eCTF}.

\pdfoutput=1
\begin{figure*}[t!]
	\centering
	\includegraphics[width=1.0\textwidth]{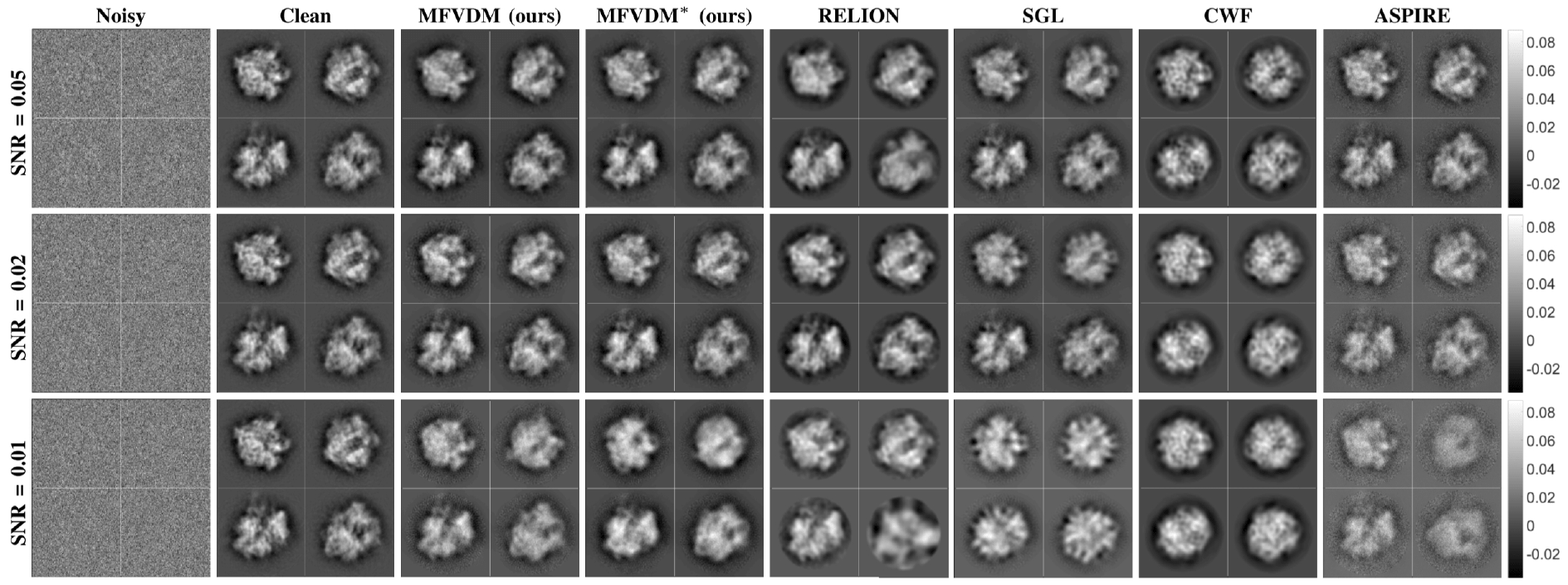}
	\caption{Image samples of simulated 70S ribosome dataset. From \textit{left} to \textit{right}: noisy projection images, clean projection images and denoised results by MFVDM, MFVDM$^*$, RELION, SGL, CWF, ASPIRE. Rows show different noise levels $\text{SNR} = 0.05,0.02,0.01$.}
	\label{fig:different_method}
\end{figure*}

\section{Experimental Results}
\label{sec:exp}
We apply our method on three cryo-EM datasets to obtain denoised images: (1) a simulated dataset %\jane{I suggest use simulated projection images. Maybe  The underlying structure is a true cryo-EM reconstruction. We just simulated the projection using the image formation model.} 
with additive white Gaussian noise and altered by CTFs, (2) two experimental datasets. All experiments were performed on a Linux system with a 60 cores Intel Xeon CPUs ($12$ cores were used), running at $2.3$GHz with $512$GB RAM in total. We compare with several state-of-the-art approaches: (1) ASPIRE\cite{zhao2014rotationally}, (2) Covariance Wiener Filter (CWF)\cite{bhamre2016denoising}, (3) RELION\cite{scheres2012relion},
(4) Steerable Graph Laplacian (SGL)\cite{landa2018steerable}.

\vspace{0.1cm}
\noindent\textbf{Parameter Setting:}  %~\footnote{We put detailed discussion in the supplementary material.}\quad}
In all experiments, for MFVDM we use $50$ nearest neighbors to build the graph, we set $\tilde{k} = 10$, $m = 50$ and denoising filter $h(\lambda) = 2\lambda - \lambda^2$. In ASPIRE we also use $50$ nearest neighbors to generate class averages, and in RELION we set $200$ classes with $25$ iterations (50 particles per class has in average). In SGL we follow the algorithms in ~\cite{landa2018steerable}, where we use $K = 256$ point FFT and keep $k_m = 50$ smallest eigenvalues for each $m$. Empirically we find these settings efficient and effective, and all these methods are not sensetive to the choice of parameters. More details are available in the appendix.%~\footnote{We provide detailed discussion in the supplementary material.}.

\subsection{Simulated Dataset - 70S ribosome}
The simulated dataset is prepared from the 3D structure of 70S ribosome, the original volume of 70S ribosome is $129 \times 129 \times 129$ voxels and the volume is centered. Then $n = 10^4$ projection images with $129 \times 129$ are generated from uniformly distributed directions over $SO(3)$. Moreover, the effects of contrast transfer functions (CTFs) are included~\footnote{Here all images are divided into $N = 20$ different defocus groups and $1\mu\mathrm{m} \leq \Delta z \leq 4\mu \mathrm{m}$. The electron beam energy $E = 200$KeV with corresponding wavelength $\lambda = 0.025$\AA~and spherical aberration $c_s = 2$mm. Pixel size is $2.82$\AA.}. In addition, white Gaussian noise is added with various noise levels as signal to noise ratio (SNR) $ = 0.05,0.02,0.01$. Notebly these are typical cases in real problem. Samples are shown in Fig.~\ref{fig:clean_noisy}. 
% The synthetic dataset is prepared from the 3D structure of 50S ribosome, the original volume of 50S ribosomal subunit is $129 \times 129 \times 129$ voxels and the volume is centered. Then $n = 10^4$ projection images with $129 \times 129$ are generated from uniformly distributed directions over $SO(3)$ (See Fig..\ref{fig:clean_noisy}(a)). Moreover, we include the effects of contrast transfer functions (CTFs) which is defined in \eqref{eq:CTF}. All images are divided into $N = 20$ different defocus groups and $1\mu\mathrm{m} \leq \Delta z \leq 4\mu \mathrm{m}$. The electron beam energy $E = 200$KeV with corresponding wavelength $\lambda = 0.025$\AA~and spherical aberration $c_s = 2$mm. Pixel size is $2.82$\AA. Examples of clean images modified by CTFs are shown in Fig.~\ref{fig:clean_noisy}(b). In addition, white Gaussian noise is added with various noise levels as signal to noise ratio (SNR) $ = 0.05,0.02,0.01$ (see Fig..\ref{fig:clean_noisy}(c) to \ref{fig:clean_noisy}(e)), such noise levels are the usual cases in real problems.
We evaluate our results on three aspects: (1) nearest neighbor search, (2) rotational alignment, and (3) denoising.

\begin{figure}[t!]
    \captionsetup[subfigure]{labelformat=empty}
    % \hspace{1.55cm}\scriptsize{\textbf{SNR = 0.05}} \hspace{4.3cm}\scriptsize{\textbf{SNR = 0.02}} \hspace{4.1cm}\scriptsize{\textbf{SNR = 0.01}}\\
    \centering
    \hspace{0.5cm}
    \subfloat{\scriptsize{\textbf{Nearest neighbor search}}}\hspace{1.7cm}
    \subfloat{\scriptsize{\textbf{Rotational alignment}}}\\[-10pt]
    \subfloat{\raisebox{0.65\height}{\rotatebox{90}{\textbf{\scriptsize{SNR = 0.05}}}}}
    \subfloat{%
    \includegraphics[width = 0.23\textwidth]{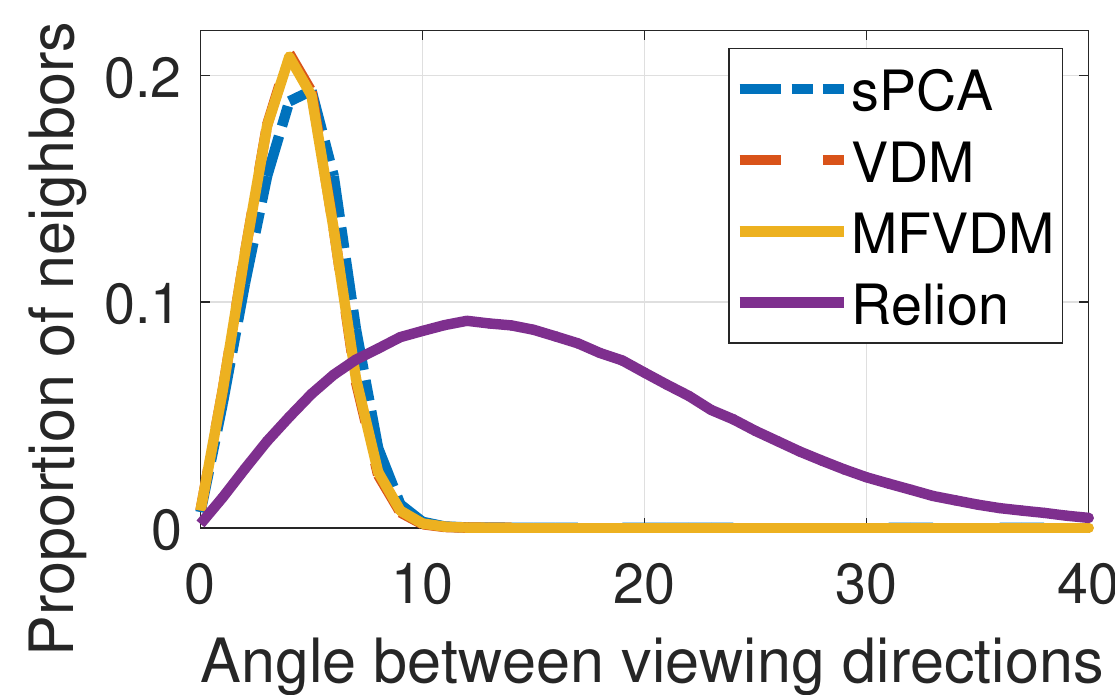}
    \label{fig:snr005}
    }
    \subfloat{%
    \includegraphics[width = 0.23\textwidth]{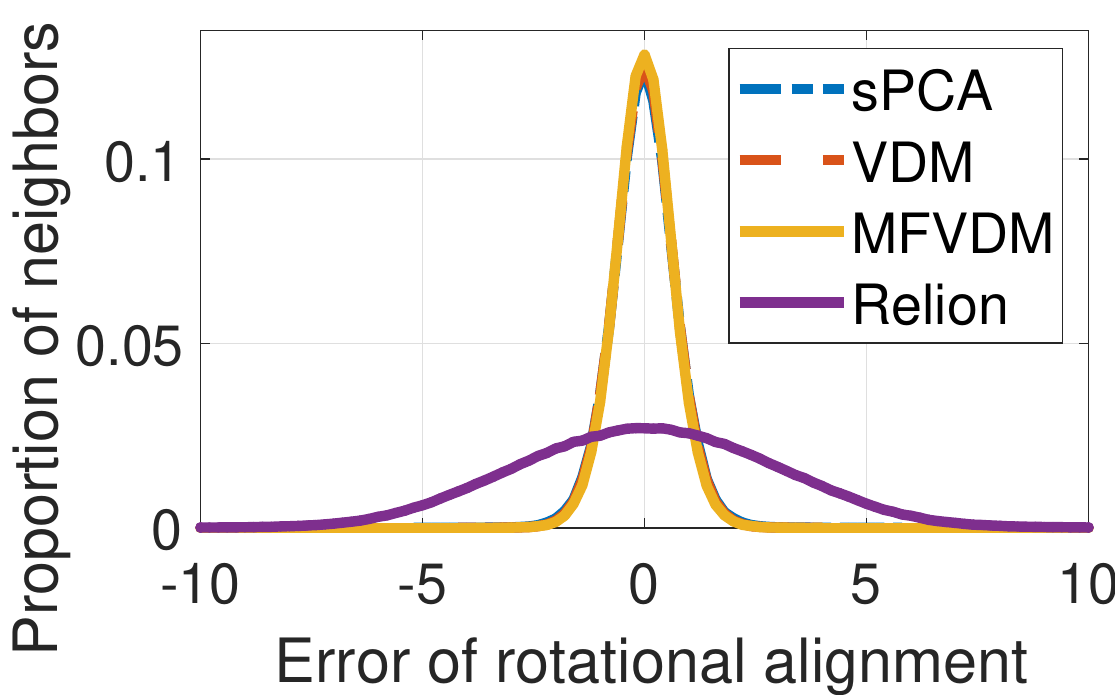}}\\
    \vspace{-0.2cm}
    \subfloat{\raisebox{0.65\height}{\rotatebox{90}{\textbf{\scriptsize{SNR = 0.02}}}}}
    \subfloat{%
    \includegraphics[width = 0.23\textwidth]{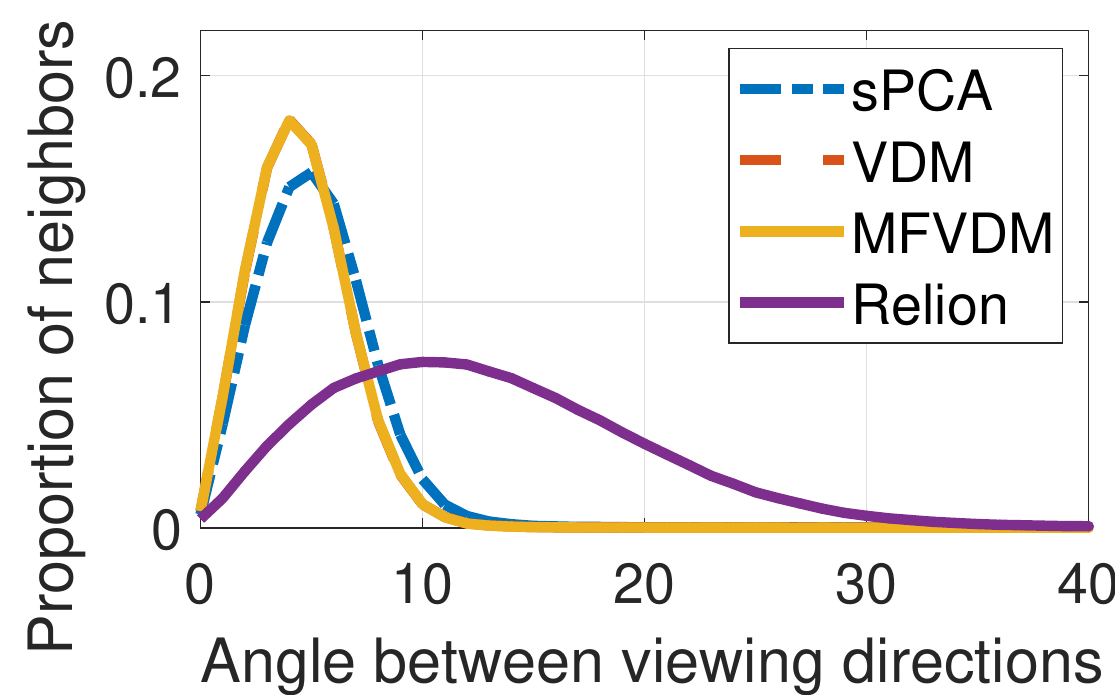}
    \label{fig:snr002}
    }
    \subfloat{%
    \includegraphics[width = 0.23\textwidth]{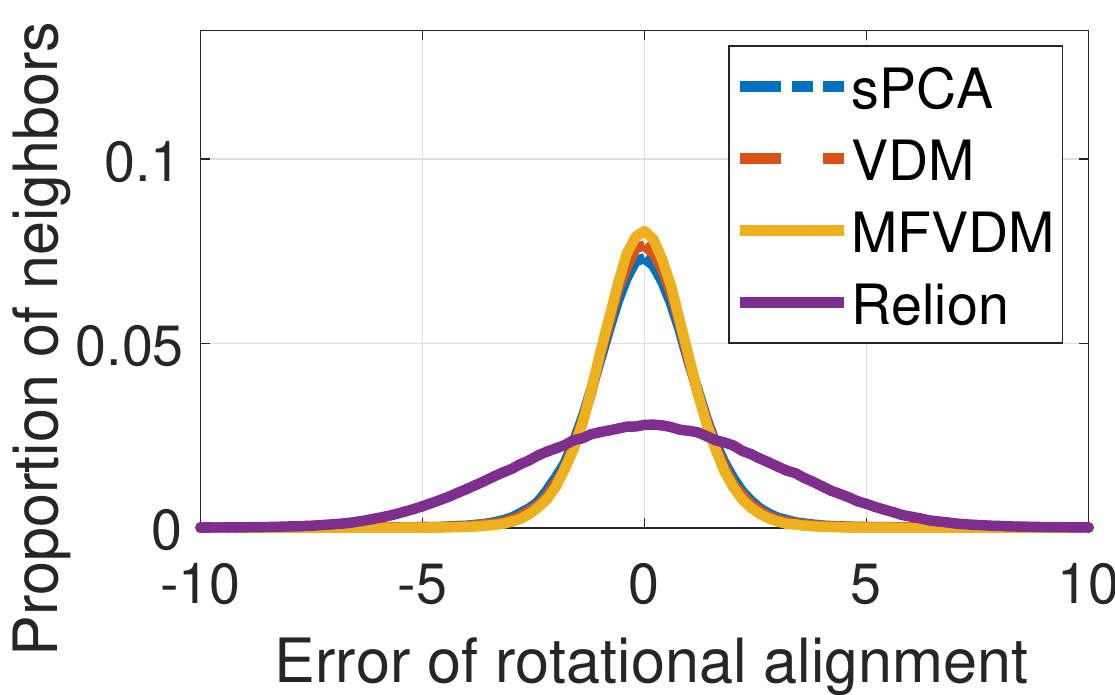}}\\
    \vspace{-0.2cm}
    \subfloat{\raisebox{0.65\height}{\rotatebox{90}{\textbf{\scriptsize{SNR = 0.01}}}}}
    \subfloat{%
    \includegraphics[width = 0.23\textwidth]{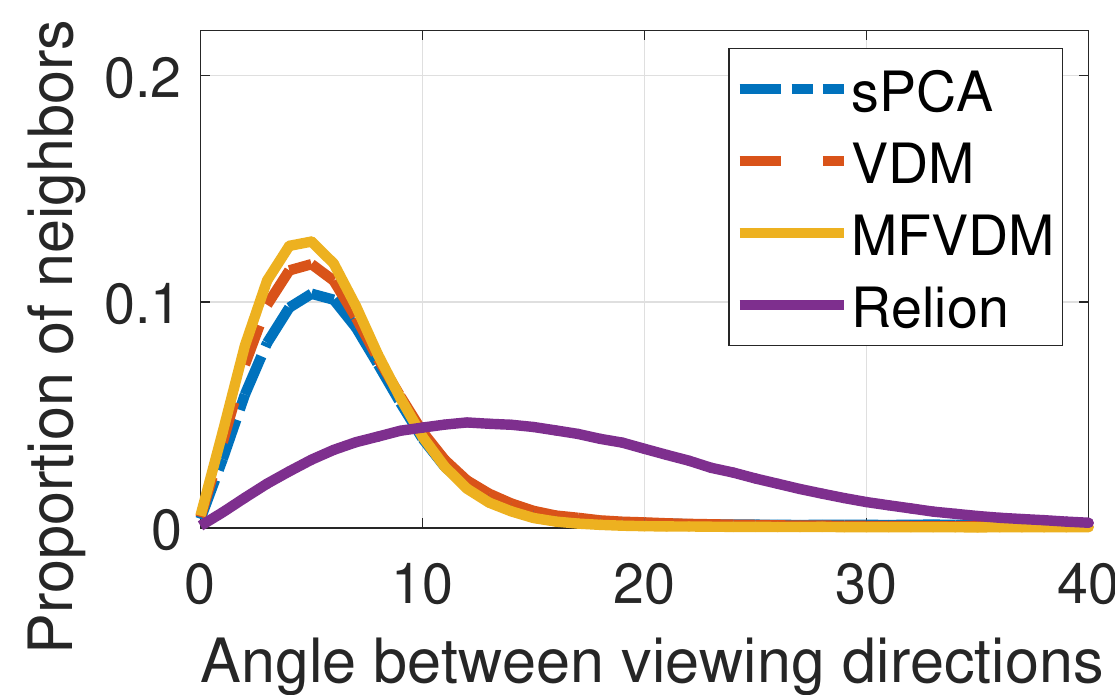}
    \label{fig:snr001}}
    \subfloat{%
    \includegraphics[width = 0.23\textwidth]{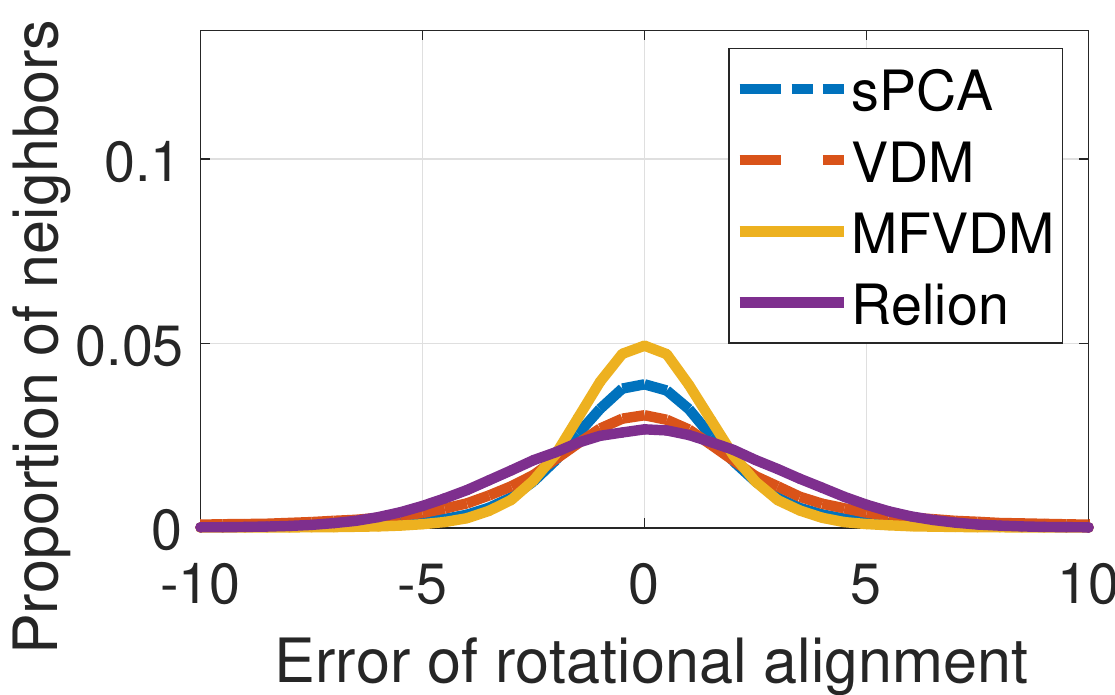}}
    \caption{Histograms of \textit{Left}: angles between viewing directions of nearest neighbors identified by different approaches, \textit{Right}: rotational alignment errors by different approaches. The horizontal axes are in degrees. Rows show different noise levels $\text{SNR} = 0.05, 0.02, 0.01$. Lines of VDM and MFVDM overlap in the top two rows.}
    \label{fig:classification_rot}
\end{figure}

\vspace{0.1cm}
\noindent\textbf{Nearest Neighbor Search: }
We evaluate nearest neighbor search by inspecting the distribution of angles between the viewing directions of nearest neighbor pairs, defined as $\theta_{ij} = \arccos \langle v_i, v_j\rangle.$
Accurate search means all $\theta_{ij}$s in the same group should concentrate to $0$ within a small spherical cap. Since the viewing direction of each projection image is known, we plot the histograms of $\theta_{ij}$ for nearest neighbor lists identified by different appoaches. Fig.~\ref{fig:classification_rot} shows, at high SNR, MFVDM and VDM have similar results. When $\text{SNR} = 0.01$, MFVDM identifies more nearest neighbors that concentrate in the small spherical cap than other methods. This indicates that MFVDM is more robust to noise.

\vspace{0.1cm}
\noindent\textbf{Rotational Alignment: }
We evaluate alignment by computing $\alpha_{ij} - \hat{\alpha}_{ij}$ for all pairs of nearest neighbors $(i,j)$, where $\alpha_{ij}$ is ground truth and $\hat{\alpha}_{ij}$ is the estimate. Fig.~\ref{fig:classification_rot} shows the histograms of errors for all nearest neighbor pairs. Note that at $\text{SNR} = 0.01$, MFVDM achieves more accurate estimation of in-plane rotational alignment than other methods, which indicates, again, MFVDM is more robust to noise. 

% First of all, let us explore how parameters of eVDM could influence the results. One uncertainty in eVDM is the choice of graph filter, here we tried four different filtering schemes in $e$VDM denoising could help: $e$VDM(1) applies $S_k$ directly to $A^{(k)}$ and $e$VDM(2) applies $2S_{k}-S_{k}^2$ directly to $A^{(k)}$. $e$VDM(3) truncates the eigenvalues with $m = 50$ and  $h(\lambda) = \lambda$; $e$VDM(4) truncates the eigenvalues with $m = 50$ and $h(\lambda) = 2\lambda - \lambda^2$. The corresponding denoised results are shown in the top row of Fig.~\ref{fig:eVDM_filter}. Each column represents a scheme. As a result, the denoised images from $e$VDM(3) and $e$VDM(4), which include the truncation of eigenvalues, looks much cleaner than the images from other two schemes. Therefore, we argue that he truncation of the eigenvalues is necessary for denoising the expansion coefficients.

\vspace{0.1cm}
\noindent\textbf{Denoising: }
Fig.~\ref{fig:different_method} presents the denoising of CTF-affected image samples at various SNR levels. To emphasize the significance of a denoised graph by MFVDM (see Fig.~\ref{fig:eVDM_illustration}), we use sPCA to build the noisy graph then denoising, which is denoted as MFVDM$^{*}$.
% In eVDM we use $50$ nearest neighbor to build the graph, we set the truncation $m = 75$ and denoising filter $h(\lambda) = 2\lambda - \lambda^2$. In ASPIRE we also use $50$ nearest neighbors to generate class averages, and in RELION we set $200$ number of classes with $25$ iterations. 
It can be seen that all methods preform well at high $\text{SNR} = 0.05$. While at low SNR cases, MFVDM and RELION outperform SGL, %~\footnote{ We added the CTF correction~\eqref{eq:CTFcorrect} to  SGL\cite{landa2018steerable} for comparisons.},
CWF and ASPIRE, whose results are blurry and have lost detailed information. In addition, MFVDM outperforms MFVDM$^{*}$, especially for $\text{SNR} = 0.01$, this indicates the effects of MFVDM neighbor search and alignment. In Appendix we show more denoised results.%\jane{You can just say for eVDM$^{*}$, we skip the step which denoises the estimated graph structure from sPCA.}

\begin{figure*}
	\centering
	\includegraphics[width=0.99\textwidth]{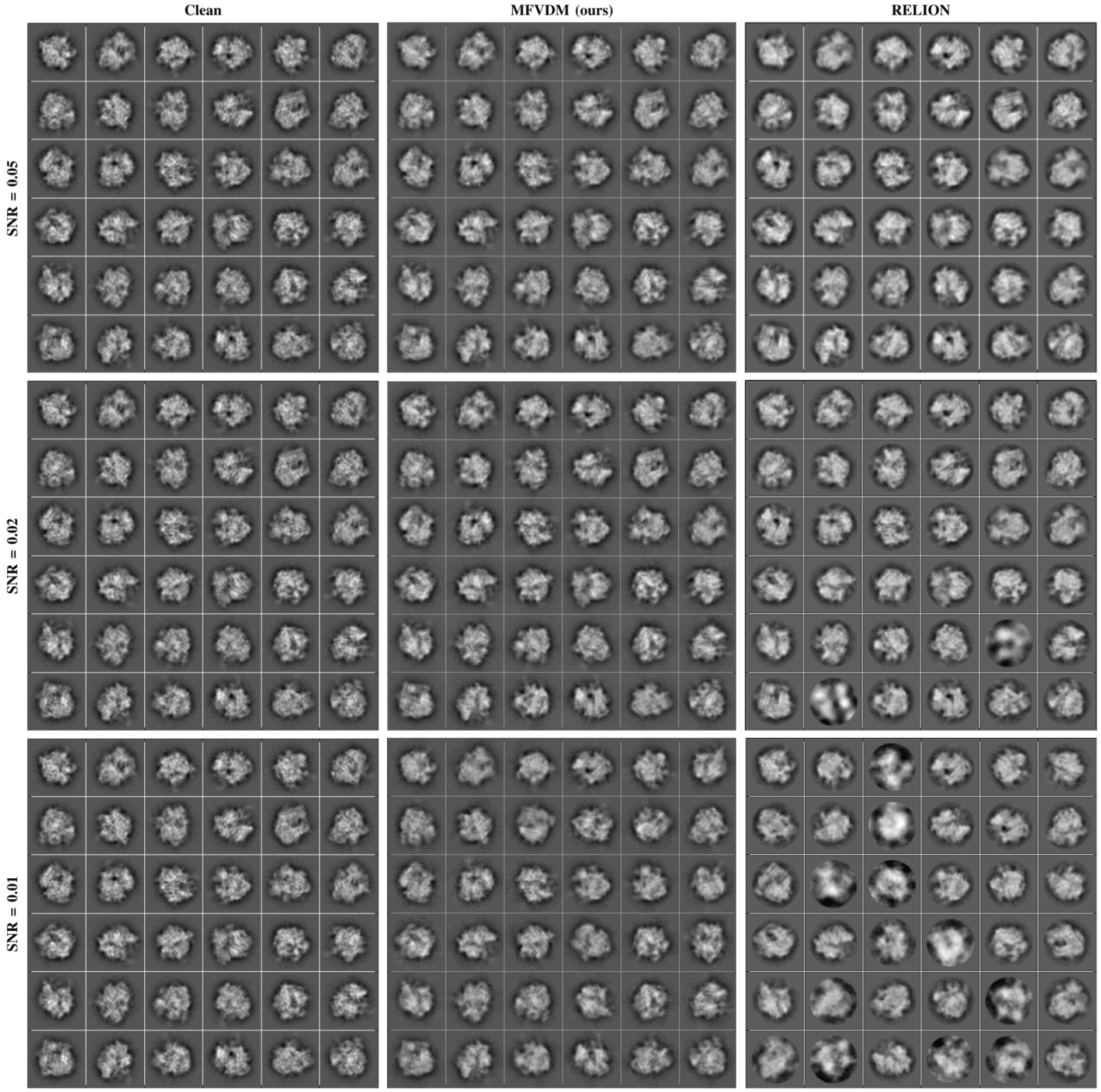}
	\caption{More sample images for denoising simulated projections from 70S ribosome, by MFVDM and RELION, at noise levels $\text{SNR} = 0.05, 0.02, 0.01$.  %\jane{Are they of the same color scale? Looks like the images in the lower corner are of different brightness scale}
	}
	\label{fig:sim_grid}
\end{figure*}

In Fig.~\ref{fig:sim_grid} we present more denoising samples to compare the results from MFVDM and RELION, at $\text{SNR} = 0.05, 0.02, 0.01$. 
It shows at high SNR, e,g, $\text{SNR} = 0.05$, both methods achieve a good performance. While at lower SNR cases as $0.02$ or $0.01$, RELION could fail to denoise on some images, which are shown in chaos. Also a few images have been classified incorrectly. However, MFVDM still almost work. Again, this show the robustness of MFVDM.

\begin{table}[t!]
% \scriptsize
\caption{Average MSE, PSNR and SSIM of denoised images on 70S ribosome. The reference are clean projection images without CTFs (see Fig.~\ref{fig:clean_noisy}(a)).}
\centering
\begin{tabular}{p{0.025\textwidth}<{\centering}| p{0.035\textwidth}<{\centering}| p{0.053\textwidth}<{\centering} p{0.053\textwidth}<{\centering} p{0.053\textwidth}<{\centering} p{0.053\textwidth}<{\centering}  p{0.053\textwidth}<{\centering}} \toprule
% \multicolumn{2}{c|}{\multirow{2}{*}{\raisebox{-0.2pt}{SNR}}} & \multicolumn{5}{c}{Method} \\ %\cmidrule{3-7}
% \multicolumn{2}{c|}{}  &MFVDM &MFVDM$^{*}$ &SGL%\cite{zhao2016fast} 
% &CWF%\cite{bhamre2016denoising}
% &ASPIRE%\cite{}
\multicolumn{2}{l|}{SNR} &MFVDM &MFVDM$^{*}$ &SGL%\cite{zhao2016fast} 
 &CWF%\cite{bhamre2016denoising}
 &ASPIRE%\cite{}
 \\ \midrule
\multirow{3}{*}{0.05} &MSE &\textbf{0.0177} &0.0186 &0.0193 &0.0313 &0.0376\\
                  &PSNR &\textbf{47.7} &47.3 &47.2 &45.2  &44.3 \\
                  &SSIM &\textbf{0.966} &0.965 &0.963 &0.943 &0.950 \\ \midrule
\multirow{3}{*}{0.02} &MSE &\textbf{0.0223} &0.0267 &0.0280 &0.0487 &0.0972 \\
                  &PSNR &\textbf{46.7} &46.0 &45.6 &43.2  &40.2 \\
                  &SSIM &\textbf{0.959} &0.953 &0.948 &0.920 &0.885 \\ \midrule
\multirow{3}{*}{0.01} &MSE &\textbf{0.0383} &0.0429 &0.0527 &0.0637 &0.273 \\
                  &PSNR &\textbf{44.6} &43.3 &43.1 &42.0  &35.6 \\
                  &SSIM &\textbf{0.932} &0.914 &0.910 &0.902 &0.715 \\ 
\bottomrule
\end{tabular}
\label{tab:mse,ssim,psnr}
\end{table}

\vspace{0.1cm}
\noindent\textbf{MSE, SSIM and PSNR: }
In Tab.~\ref{tab:mse,ssim,psnr} we present the average of mean squared error (MSE), peak SNR (PSNR) and structural similarity index (SSIM) on the denoised images. Notebly MFVDM outperforms other approaches with the smallest MSE and largest PSNR and SSIM, at all SNR levels. For RELION, since its denoising result includes unknown inplane shifts, we are unable to fairly evaluate such measurements on RELION.

\begin{figure*}[t!]
	\centering
	\footnotesize
	\vspace{0.4cm}
	\includegraphics[width=0.85\textwidth]{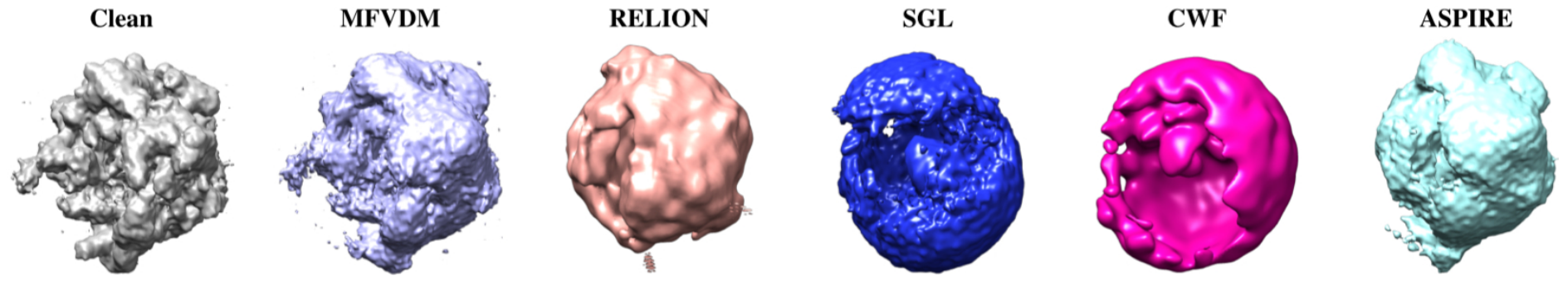}
	
	\caption{3D reconstructions of 70S ribosome using 1000 denoised images as shown in Fig.~\ref{fig:different_method} at $\text{SNR} = 0.01$. %, from denoising results by eVDM, RELION, SGL, CWF, and ASPIRE 
	}
	\label{fig:3D_recon}
\end{figure*}

%\begin{figure*}[t!]
%\centering
%\footnotesize
%\vspace{0.4cm}
%\begin{tabular}{p{0.12\textwidth}<{\centering} p{0.12\textwidth}<{\centering} p{0.12\textwidth}<{\centering} p{0.12\textwidth}<{\centering} p{0.12\textwidth}<{\centering} 
%p{0.12\textwidth}<{\centering} 
%} 
% \textbf{\scriptsize{Clean}} &\textbf{\scriptsize{MFVDM}}
% &\textbf{\scriptsize{RELION}}
% &\textbf{\scriptsize{SGL}} &\textbf{\scriptsize{CWF}} &\textbf{\scriptsize{ASPIRE}}\\ 
%
%\includegraphics[width = 0.12\textwidth]{figures/sim_data/denoising/70S_ref.png}
%&\includegraphics[width = 0.12\textwidth]{figures/sim_data/denoising/70S_recon_eVDM.png}
%&\includegraphics[width = 0.105\textwidth]{figures/sim_data/denoising/70S_relion200.png}
%&\includegraphics[width = 0.105\textwidth]{figures/sim_data/denoising/70S_nsr100_SGL.png}
%&\includegraphics[width = 0.11\textwidth]{figures/sim_data/denoising/70S_recon_CWF2.png}
%&\includegraphics[width = 0.10\textwidth]{figures/sim_data/denoising/70S_recon_VDM.png}\\
%\end{tabular}
%
%\caption{3D reconstructions of 70S ribosome using 1000 denoised images as shown in Fig.~\ref{fig:different_method} at $\text{SNR} = 0.01$. %, from denoising results by eVDM, RELION, SGL, CWF, and ASPIRE 
%}
%\label{fig:3D_recon}
%\end{figure*}

\vspace{0.1cm}
\noindent\textbf{Runtime: }
Tab.~\ref{tab:runtime} compares the runtime of different approaches. We take $n = 10^4$ noisy images with $\text{SNR} = 0.01$ as input, and output all denoised ones. All methods use the same amount of computational resource (12 cores CPU). 
% Both ASPIRE and MFVDM use sPCA for initial classification, and for CWF and RELION we only present the total runtime due to their different procedures. 
The MFVDM denosing, by building a nearest neighbor graph and filtering it, runs much faster than RELION 
which is an iterative procedure. Even though with GPU implementation, RELION still takes more time. The runtime of SGL and ASPIRE are similar to MFVDM due to their similar procedures. This demonstrates that MFVDM is more efficient compared to other approches.

% \setlength{\textfloatsep}{0.2cm}
% \begin{table}[t!]
% \scriptsize
% \caption{The runtime (in seconds) of different methods on 70s ribosome, at $\text{SNR} = 0.01$. ASPIRE and MFVDM can be divided into 4 steps. \raisebox{.5pt}{\textcircled{\raisebox{-.9pt} {1}}}: sPCA.  \raisebox{.5pt}{\textcircled{\raisebox{-.9pt} {2}}}: initial neighbor search and rotational alignment. \raisebox{.5pt}{\textcircled{\raisebox{-.9pt} {3}}}: further neighbor search and rotational alignment. \raisebox{.5pt}{\textcircled{\raisebox{-.9pt} {4}}}: denoising.
%     }
% \centering
% \vspace{-0.2cm}
% \begin{tabular}{p{0.05\textwidth}<{\centering}| p{0.05\textwidth}<{\centering} p{0.07\textwidth}<{\centering} p{0.05\textwidth}<{\centering}  p{0.05\textwidth}<{\centering} p{0.05\textwidth}<{\centering}} \toprule
% % \multirow{2}{*}{Runtime} & \multicolumn{5}{c}{Method} \\ %\cmidrule{2-6}
% %  &MFVDM &RELION &SGL &CWF &ASPIRE \\ \midrule
% Runtime &MFVDM &RELION &SGL &CWF &ASPIRE \\ \midrule
% \raisebox{.5pt}{\textcircled{\raisebox{-.9pt} {1}}}(s)  &44.1 &- &44.1 &- &44.1\\
% \raisebox{.5pt}{\textcircled{\raisebox{-.9pt} {2}}}(s)   &22.1 &- &-  &- &22.1\\
% \raisebox{.5pt}{\textcircled{\raisebox{-.9pt} {3}}}(s)   &232.5 &- &- &- &41.2\\ 
% \raisebox{.5pt}{\textcircled{\raisebox{-.9pt} {4}}}(s)   &432.1 &465 mins &635.7 &647.6 &812.2\\
% Total(s)                                                 &730.8 &465 mins &679.8 &647.6 &919.6\\
% \bottomrule
% \end{tabular}
% \label{tab:runtime}
% \end{table}

\setlength{\textfloatsep}{0.2cm}
\begin{table}[t!]
% \scriptsize
\caption{The runtime (in minutes) of different methods on 70s ribosome, at $\text{SNR} = 0.01$. RELION$^{*}$ represents RELION with GPU acceleration.}
\centering
\begin{tabular}{p{0.05\textwidth}<{\centering} p{0.07\textwidth}<{\centering} p{0.07\textwidth}<{\centering} p{0.05\textwidth}<{\centering}  p{0.05\textwidth}<{\centering} p{0.05\textwidth}<{\centering}} \toprule
MFVDM &RELION &RELION$^{*}$ &SGL &CWF &ASPIRE \\ \midrule  
12 & 465 &117 & 12  & 11 & 15 \\
%730.8s &465 mins &117 mins &679.8s &647.6s &919.6s\\
\bottomrule
\end{tabular}
\label{tab:runtime}
\end{table}

\begin{table}[t!]
% \scriptsize
\caption{The volume resolutions (lower is better) for 70S ribosome }
\centering
\begin{tabular}{p{0.06\textwidth}<{\centering} p{0.06\textwidth}<{\centering}  p{0.06\textwidth}<{\centering} p{0.06\textwidth}<{\centering} p{0.06\textwidth}<{\centering}} \toprule
% \multirow{2}{*}{Dataset} & \multicolumn{6}{c}{Method} \\ %\cmidrule{2-6}
%  &MFVDM  &RELION &RELION$^{*}$ &CWF &SGL &ASPIRE\\ \midrule
MFVDM  &RELION &CWF &SGL &ASPIRE\\ \midrule
\textbf{14.8\AA} &25.4\AA  &44.4\AA &39.5\AA  &17.7\AA \\
\bottomrule
\end{tabular}
\label{tab:sim_resolution}
\end{table}

\begin{figure*}[t!]
	\centering
	\includegraphics[width=0.97\textwidth]{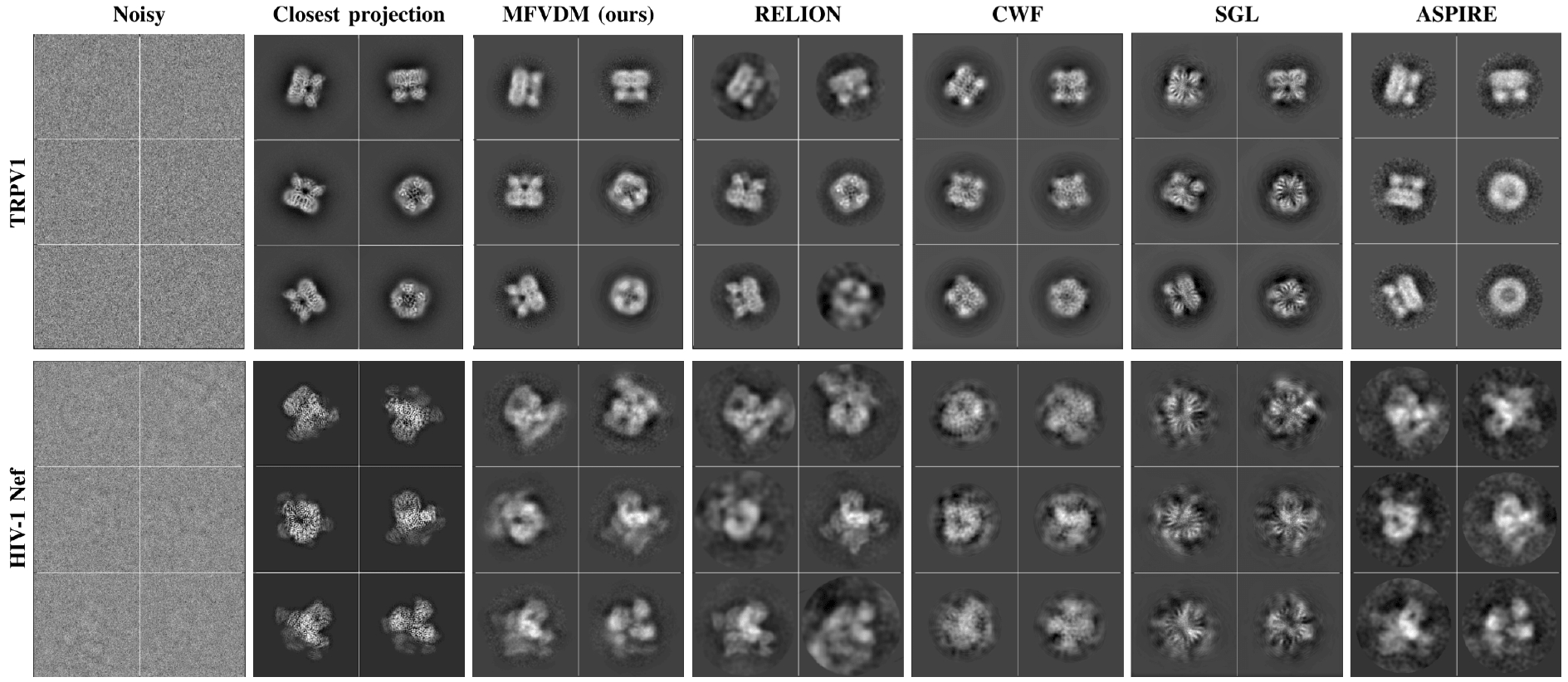}
	\caption{Image samples of \textit{top}: TRPV1; \textit{Bottom}: HIV-1 Nef datasets. From \textit{left} to \textit{right}: raw images, \textbf{closest projection: } the clean projection images from the reference map that have the highest correlation with the noisy images estimated by RELION. the denoised results by MFVDM, RELION, CWF, SGL and ASPIRE.}
	\label{fig:real_different_method}
\end{figure*}

\vspace{0.1cm}
\noindent\textbf{3D \textit{Ab Initio} Model: }
The 3D \textit{ab initio} modeling benefits from the improved denoising of 2D projection images. In Fig.~\ref{fig:3D_recon}, for each denoising method, we show the 3D reconstruction of 70S ribosome from 1000 denoised projection images (original $\text{SNR} = 0.01$) using common-lines based algorithm~\cite{wang2013exact}. The corresponding noisy images are the same for each method. It shows that only MFVDM achieves a comparatively accurate reconstruction and the results from other methods are noisy or lose significant sub-structures. In addition to  visual inspection, we use the Fourier shell correlation (FSC)~\cite{frank2006three} between the reconstruction and the original volume to quantify the resolution, which is determined by the 0.143 cutoff of FSC (see Tab.~\ref{tab:sim_resolution}).

\subsection{Experimental Datasets}
We test the performance of MFVDM on two public experimental datasets %~\footnote{Both datasets are available in the Electron Microscope Pilot Image Archive (EMPIRE) as EMPIAR-10015 (TRPV1) and EMPIAR-10178 (HIV-1 Nef).},
available in the Electron Microscope Pilot Image Archive (EMPIAR)~\cite{LudinEtAl16}: 
(1) TRPV1 ion channel\cite{liao2013structure} (for brevity, TRPV1), (2) HIV-1 Nef\cite{morris2018hiv}.
Noisy samples are shown in the first column of  Fig.~\ref{fig:real_different_method}. For each dataset we take $n = 10^4$ images as input to perform each approach. 
% Consider the experimental images are typically corrupted by colored noise, we first prewhiten the raw images, and recolor the denoised images (detailed in supplementary material). MFVDM denoising is the most efficient pipeline compared with other approaches (see Tab.~\ref{tab:real_runtime}).

\vspace{0.1cm}
\noindent\textbf{Colored Noise Whitening: }
Consider the experimental cryo-EM images are typically corrupted by colored noise,  we need to first prewhiten the raw images: We estimate the power spectrum of noise based on the corner pixel values, since particles are contained in the middle of the images. Then we whiten each image using the estimated noise power spectrum and perform CTF correction via phase flipping. We perform nearest neighbor search, alignment, and denoising on the pre-processed images. After denoising, we recolor the denoised images by multiplying the estimated noise power spectrum in the Fourier domain and perform CTF correction based on~\eqref{eq:CTFcorrect}.

\vspace{0.1cm}
\noindent\textbf{Denoising: } 
%In Fig.~\ref{fig:real_different_method} we present the denoising images. Due to the lack of ground truth images, it is impossible to compute the MSE, SSIM and PSNR. But visually, MFVDM and RELION results are cleaner than the results of other methods.
The sample denoised images are displayed in Fig.~\ref{fig:real_different_method}. Here we generated the clean template images from the reference map and estimated the most similar template for each raw image. We used the 3D auto refine tool in RELION 2.1 to carry out the template matching (notice the closest projection may not be the truth beneath the raw image due to the high level of noise). Visually, MFVDM denoised images are cleaner than the results of other methods. Due to the lack of ground truths, i.e., particle orientations and clean images, we cannot compare the classification and alignment results as before, and provide the MSEs, SSIMs and PSNRs for the denoised images. To quantitatively evaluate the methods for 2D image analysis, we evaluate the 3D \textit{ab initio} models constructed from those denoised images. In Appendix we show more denoised results.

\vspace{0.1cm}
\noindent\textbf{Runtime: } 
In Tab.~\ref{tab:real_runtime} we report the runtime of different approaches, for denoising $n = 10^4$ images. Similar to the synthetic example, MFVDM denoising is the most efficient pipeline compared with other approaches.

\begin{table}[t!]
% \scriptsize
%\vspace{-0.25cm}
\caption{The runtime (in minutes) of different methods on TRPV1 and HIV-1 Nef datasets. RELION$^{*}$ represents RELION with GPU acceleration.%\jane{Do you need to mention that RELION is run on GPU? What is the CPU time? For other two methods, we can potentially speed up by GPU as well.}
}
\centering
\begin{tabular}{p{0.07\textwidth}<{\centering}| p{0.05\textwidth}<{\centering} p{0.05\textwidth}<{\centering} p{0.05\textwidth}<{\centering} p{0.03\textwidth}<{\centering} p{0.03\textwidth}<{\centering} p{0.045\textwidth}<{\centering}} \toprule
% \multirow{2}{*}{Dataset} & \multicolumn{6}{c}{Method} \\ %\cmidrule{2-6}
%  &MFVDM  &RELION &RELION$^{*}$ &CWF &SGL &ASPIRE\\ \midrule
Dataset &MFVDM  &RELION &RELION$^{*}$ &CWF &SGL &ASPIRE\\ \midrule
TRPV1  &43 &366  &284  &82  &51  &63 \\
HIV-1 Nef &32   &308  &127  &71  &36  &49 \\
\bottomrule
\end{tabular}
\label{tab:real_runtime}
\end{table}

\begin{figure*}[t!]
    \centering
    \setlength\tabcolsep{0.1pt}
    \renewcommand{\arraystretch}{0.6}
    \begin{tabular}{p{0.02\textwidth}<{\centering} p{0.32\textwidth}<{\centering} p{0.32\textwidth}<{\centering} p{0.32\textwidth}<{\centering}}
    % \textbf{\scriptsize{View 1}} &\textbf{\scriptsize{View 2}}\\
    \raisebox{1.78\height}{\rotatebox{90}{\textbf{\scriptsize{TRPV1}}}}
    &\includegraphics[width = 0.32\textwidth]{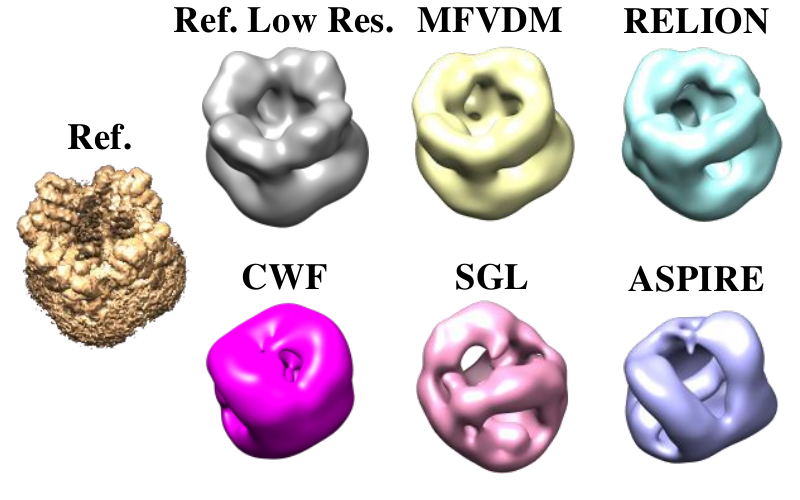}\hfill
    &\includegraphics[width = 0.32\textwidth]{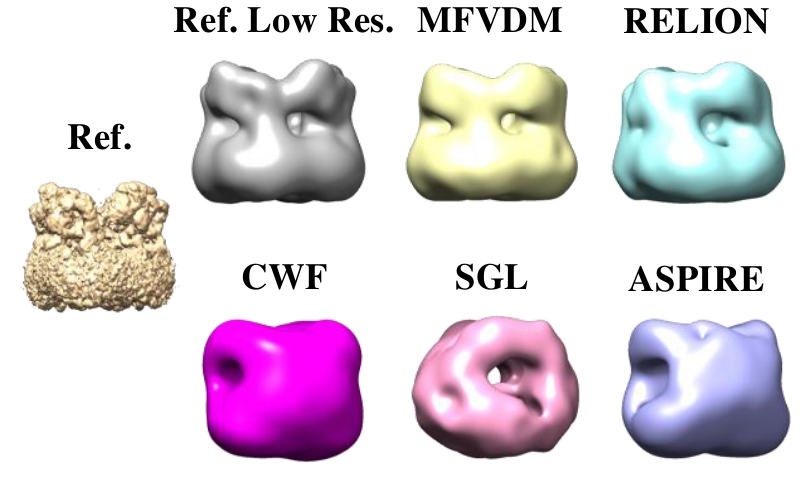}\hfill
    &\includegraphics[width = 0.32\textwidth]{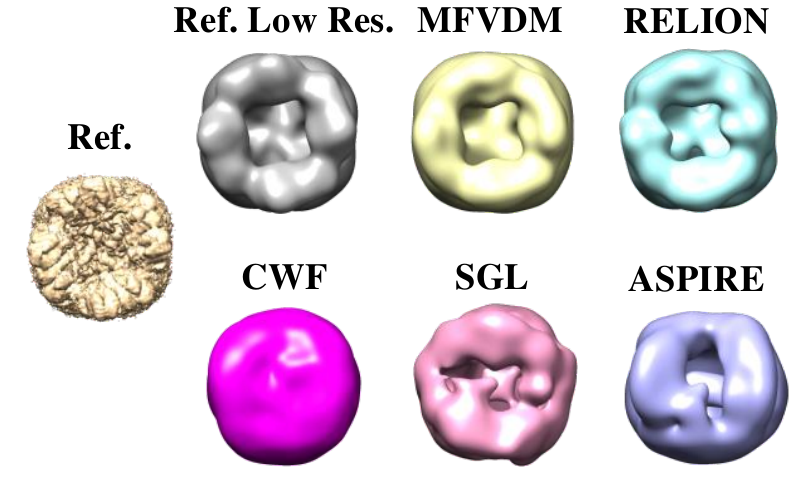}\\
    \raisebox{1.05\height}{\rotatebox{90}{\textbf{\scriptsize{HIV-1 Nef}}}}
    &\includegraphics[width = 0.32\textwidth]{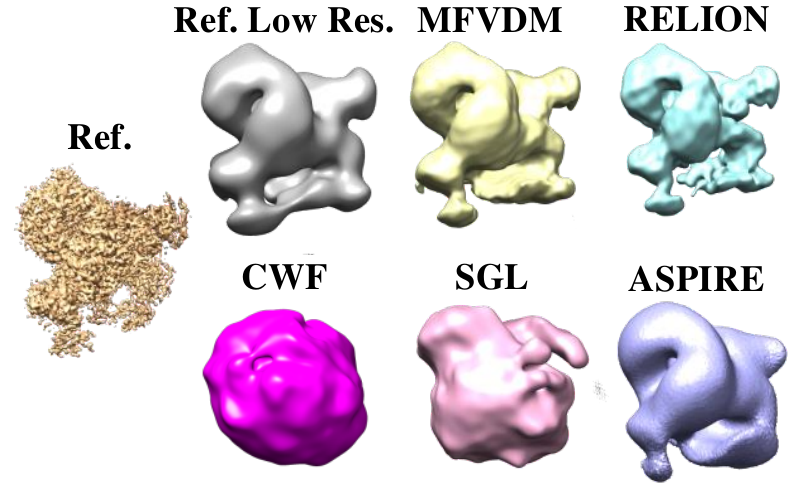}\hfill
    &\includegraphics[width = 0.32\textwidth]{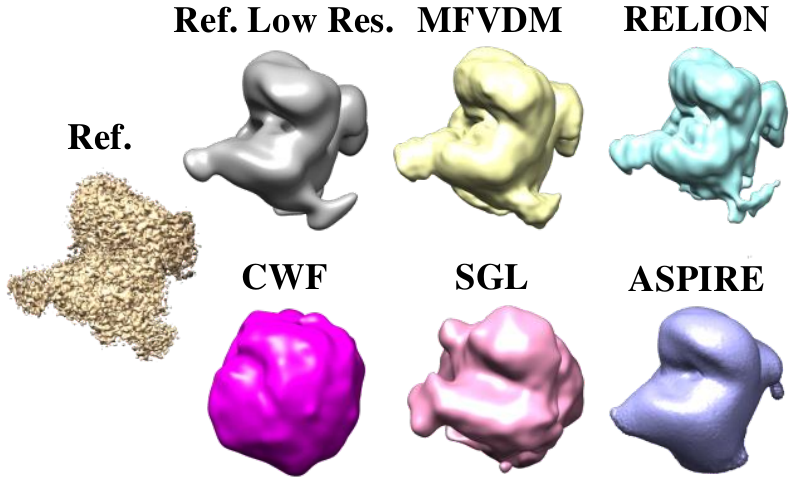}\hfill
    &\includegraphics[width = 0.32\textwidth]{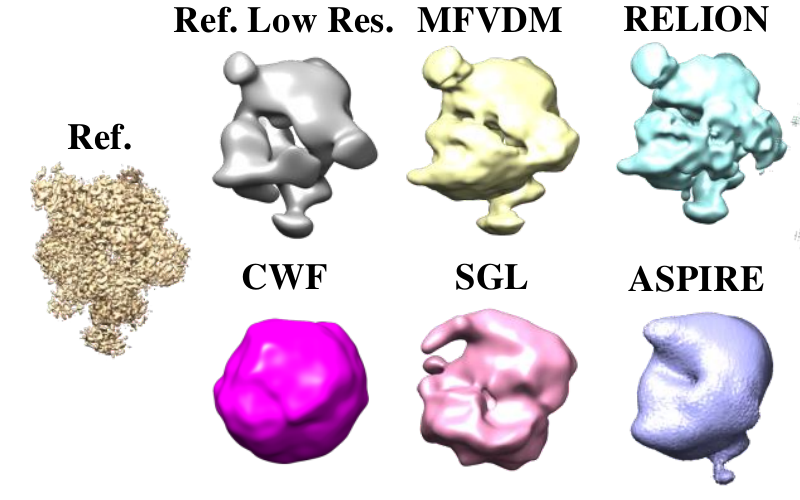}

    \end{tabular}
    \caption{3D \textit{ab initio} models of \textit{top}: TRPV1; \textit{Bottom}: HIV-1 Nef, with denoised images from MFVDM, RELION 2D class averaging, CWF, SGL, and ASPIRE 2D class averaging. Three views of the volumes are shown in columns. 'Ref.' represents the reference volume, `Ref. Low Res.' represents the Gaussian filtered reference volume. }
    \label{fig:real_recon_EM10005}
\end{figure*}

\begin{figure}[t!]
    \centering
    \subfloat{\includegraphics[width = 0.245\textwidth]{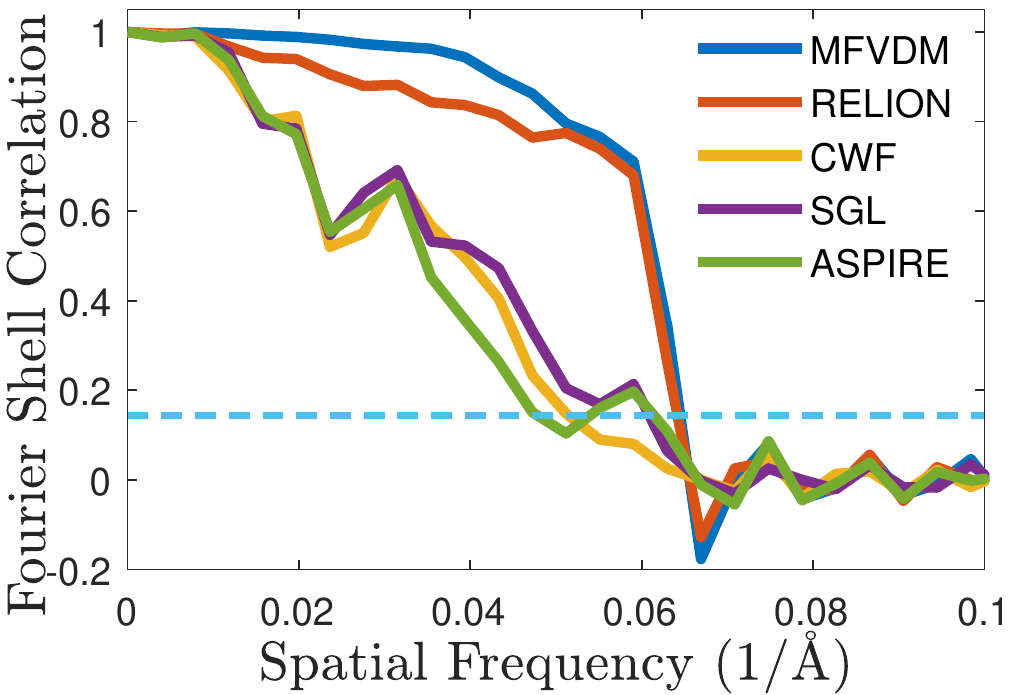}}
    \subfloat{\includegraphics[width = 0.245\textwidth]{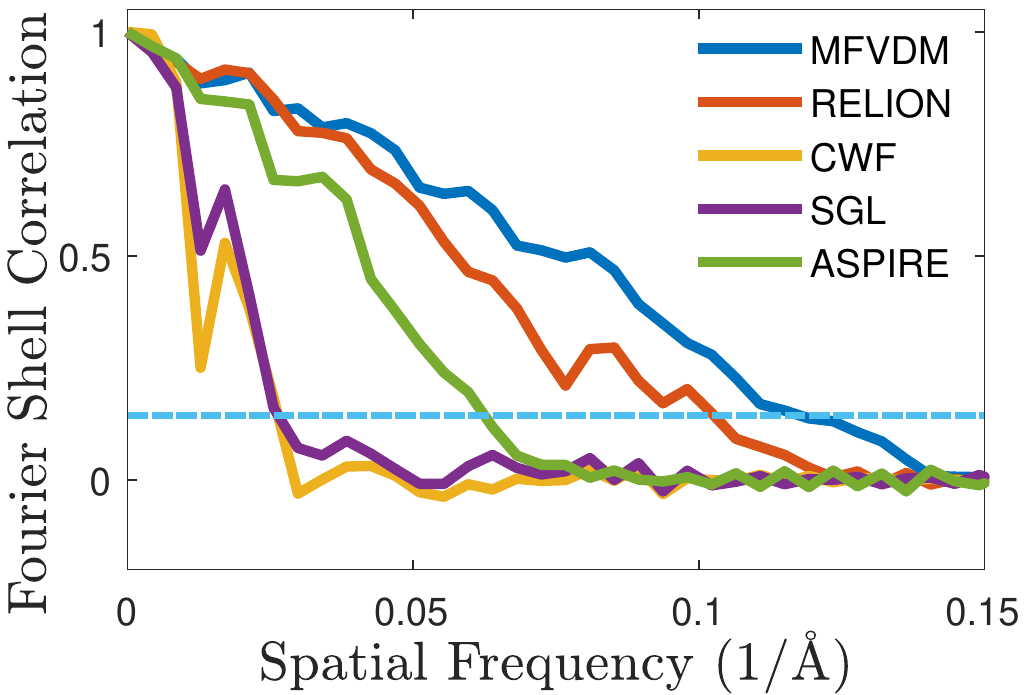}}
    \caption{Fourier shell correlation (FSC) between the reconstruction and the Reference volume (high resolution). Blue dashed line is the 0.143 cut-off criterion. \textit{Left}: TRPV1, \textit{Right}: HIV-1 Nef.}
    \label{fig:FSC_TRPV_HIV}
\end{figure}

\vspace{0.1cm}
\noindent\textbf{3D \textit{Ab Initio} Model: } In Fig.~\ref{fig:real_recon_EM10005}, we show the 3D \textit{ab initio} models generated from 1000 denoised images with identical underlying noisy images. The TRPV1 volumes are generated by \emph{RELION 3D Initial Model} with 50 iterations. The HIV-1 Nef volumes are generated by common-lines algorithm~\cite{wang2013exact}. Fig.~\ref{fig:real_recon_EM10005} shows that MFVDM and RELION are able to achieve good initial models that are similar to the Gaussian filtered reference volumes. Other methods lose significant sub-sructure. In addition, we compute the Fourier shell correlation (FSC) in Fig.~\ref{fig:FSC_TRPV_HIV}. It shows that the reconstruction based on the MFVDM denoised images achieves the highest correlations and resolutions (see Tab.~\ref{tab:real_resolution}) for both datasets. Our new pipeline outperforms the state-of-the-art methods and is able to efficiently produce accurate initial models for cryo-EM.  %Moreover, in Tab.~\ref{tab:real_resolution} we show the resolutions determined by the 0.143 cutoff of FSCs (blue dash line in Fig.~\ref{fig:FSC_TRPV_HIV}). 

\begin{table}[t!]
% \scriptsize
\caption{The volume resolutions (lower is better) for TRPV1 and HIV-1 Nef.}
\centering
\begin{tabular}{p{0.07\textwidth}<{\centering}| p{0.055\textwidth}<{\centering} p{0.055\textwidth}<{\centering}  p{0.05\textwidth}<{\centering} p{0.05\textwidth}<{\centering} p{0.05\textwidth}<{\centering}} \toprule
% \multirow{2}{*}{Dataset} & \multicolumn{6}{c}{Method} \\ %\cmidrule{2-6}
%  &MFVDM  &RELION &RELION$^{*}$ &CWF &SGL &ASPIRE\\ \midrule
Dataset &MFVDM  &RELION &CWF &SGL &ASPIRE\\ \midrule
TRPV1  &\textbf{15.88\AA} &\textbf{15.88\AA}  &16.93\AA &19.54\AA  &21.16\AA \\
HIV-1 Nef &\textbf{8.69\AA} &9.78\AA &39.12\AA  &39.12\AA  &16.76\AA \\
\bottomrule
\end{tabular}
\label{tab:real_resolution}
\end{table}

% \begin{figure*}[htb!]
%     \centering
%     \setlength\tabcolsep{0.1pt}
%     \renewcommand{\arraystretch}{0.6}
%     \begin{tabular}{p{0.33\textwidth}<{\centering} p{0.33\textwidth}<{\centering} p{0.33\textwidth}<{\centering}}
%     % \textbf{\scriptsize{View 1}} &\textbf{\scriptsize{View 2}}\\

%     \end{tabular}
%     \caption{3D \textit{ab initio} models of HIV-1 Nef with denoised images from MFVDM, RELION 2D class averaging, CWF, SGL, and ASPIRE 2D class averaging. Two views of the volumes are shown. }
%     \label{fig:real_recon_EM10178}
% \end{figure*}

% The resolutions for the reconstructed TRPV1 are: \textbf{15.88\AA (MFVDM, this paper)}, 15.88\AA (Relion),  19.54\AA (CWF),  16.93\AA (CWF), and 21.16\AA (ASPIRE).  

% The resolutions for the reconstructed HIV-1 Nef are: \textbf{8.69\AA (MFVDM, this paper)}, 9.78\AA (Relion),  39.12\AA (CWF),  39.12\AA (CWF), and 16.76\AA (ASPIRE). 
%\frank{Some part that need a discusstion: 1.Shift problem. 2.How to handle larger dataset efficiently.3.Parameters}

% !TEX root = main.tex

\section{Conclusion}
\label{sec:concl}
%\frank{Can we simplify this part? } \jane{What do you want to remove here?} \frank{Just feel the conclusion part is too long.}
%\frank{Change the Conlusion to Discussion and Conclution?}
Cryo-EM imaging will benefit from the cross fertilization of ideas with the computer vision community. %In this paper, we  a framework for recovering the single particle %\frank{Also I think we may not need to emphasize too much on our methods here.} \jane{need a short summary of what we did} 
Here, we proposed a new framework for cryo-EM image restoration by defining a new metric based on MFVDM to improve the rotationally invariant nearest neighbor search among a large number of extremely noisy images. This new metric takes into account of the consistency of the irreducible representations of the alignments along connected paths. The eigenvectors of the MFVDM matrices contain alignment information for images of similar views and are therefore used for optimal alignment estimation. %and we developed an FFT based approach for using the eVDM eigenvectors to estimate the optimal alignments. 
The denoised graph structure allows the computation of robust global basis on the data manifold. The resulting graph filtering schemes are able to efficiently denoise the Fourier-Bessel expansion coefficients of the images and perform CTF correction. %The denoised images can be efficiently synthesized from the expansion coefficients. 
%The new pipeline is very efficient, because it is based on the computation of top eigenvectors of several sparse Hermitian matrices that runs in parallel. 
%Experiment.
Through the simulated and experimental datasets, we demonstrated our proposed methods outperform the state-of-the-art 2D class averaging and image restoration algorithms. %Some methods introduced here can be applied to camera location estimation for 3D object reconstruction. 
%To that end, 
%\frank{altough we have shown the robustness of our approach on experimental datasets, it could be helpful to take into account the effects of inplane shifts in the future work. Through this paper we hope to make cryo-EM problems accessible to most computer vision audiences. }

% % keywords can be removed
% \keywords{First keyword \and Second keyword \and More}
% \clearpage
{\small
\bibliographystyle{ieee}
\bibliography{ref_denoise,refeccv}
}

\appendix
\pdfoutput=1
\subsection{More Experimental Results}
\label{sec:exp}
In this section, we provide more experimental results on the simulated dataset (70S ribosome) and the experimental datasets (TRPV1 and HIV-1 Nef). The parameter setting is the same as the main paper.

\begin{figure*}[t!]
	\centering
	\includegraphics[width=0.98\textwidth]{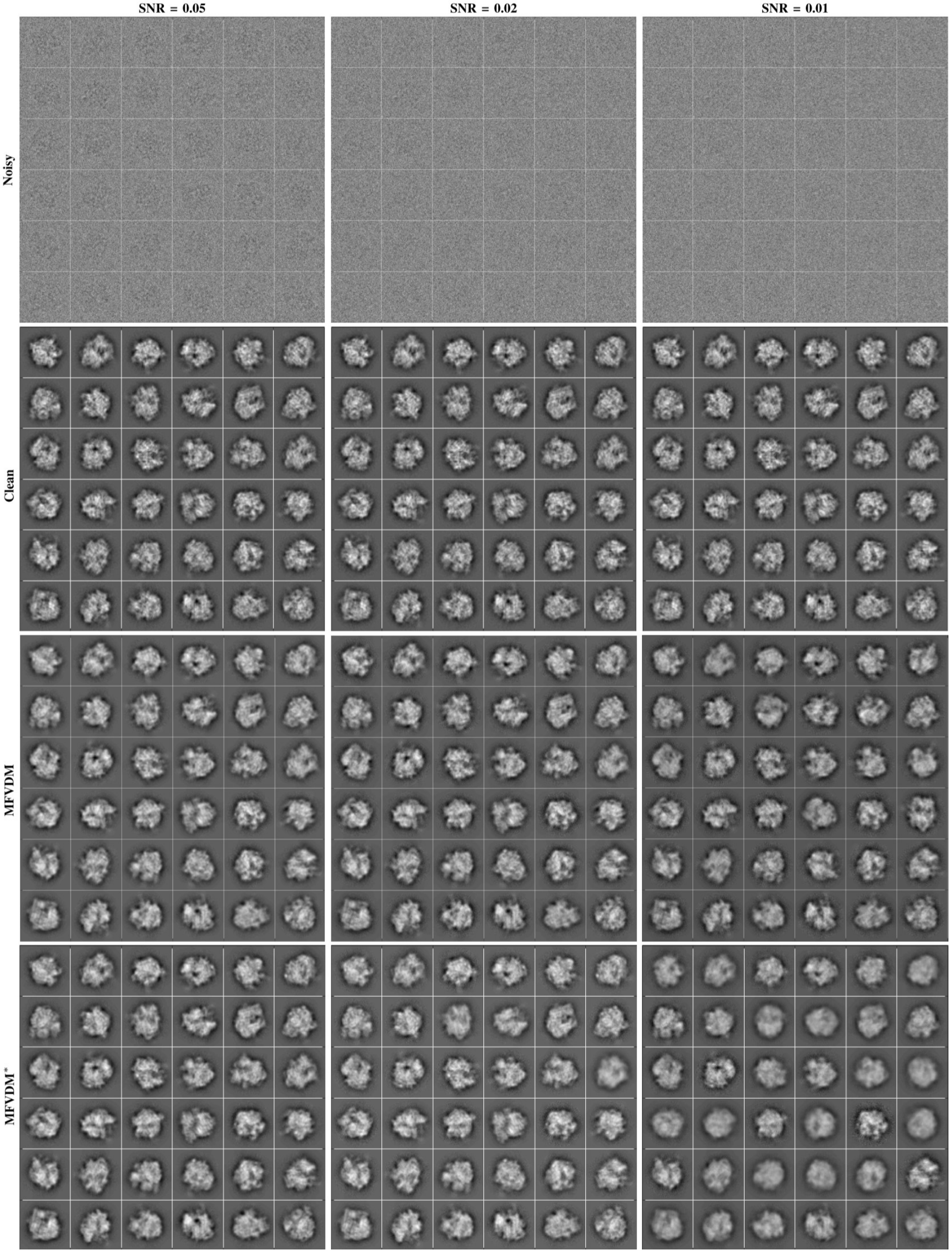}
	\caption{Image samples of simulated 70S ribosome dataset. From \textit{top} to \textit{bottom}: noisy projection images, clean projection images and denoised results by MFVDM, MFVDM$^*$. Columns show different noise levels $\text{SNR} = 0.05,0.02,0.01$.}
	\label{fig:sim_denoise_1}
\end{figure*}

\begin{figure*}[t!]
	\centering
	\includegraphics[width=0.98\textwidth]{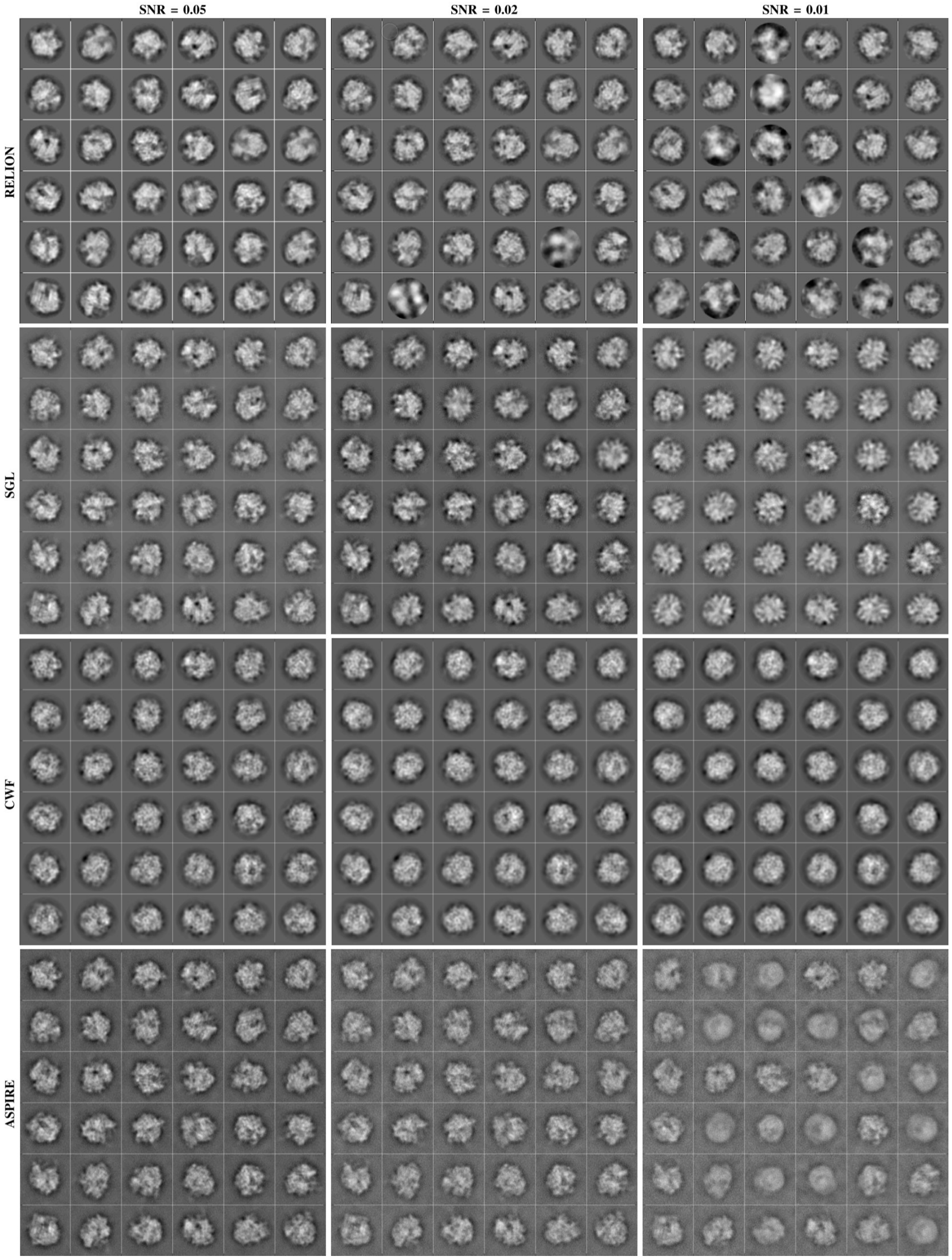}
	\caption{Image samples of simulated 70S ribosome dataset. From \textit{top} to \textit{bottom}: denoised results by RELION, SGL, CWF and ASPIRE. Columns show different noise levels $\text{SNR} = 0.05,0.02,0.01$.}
	\label{fig:sim_denoise_2}
\end{figure*}

\vspace{0.1cm}
\noindent\textbf{Simulated Dataset: }In Fig.~\ref{fig:sim_denoise_1} and Fig.~\ref{fig:sim_denoise_2} we show more denoised images of 70S ribosome by MFVDM, MFVDM$^*$, RELION 2D class averaging, SGL, CWF, and ASPIRE 2D class averaging at SNR $= 0.05, 0.02, 0.01$. All methods preform well at high SNR ($\text{SNR} = 0.05$). As the noise level increases, MFVDM outperforms other methods. Some class averages generated from RELION does not have any structural information and a few images are classified incorrectly. SGL and CWF are able to denoise and preserve the main structure, but some detailed information has been lost. ASPIRE appears noisier due to the fact that it only averages $s$-nearest neighbors directly. It shows that compared with the existing methods, MFVDM is more robust to noise. In addition, the performance of MFVDM$^*$ (without Alg. 1 in the main paper) is worse than MFVDM, especially at $\text{SNR} = 0.01$, this indicates the importance of MFVDM nearest neighbor search and alignment. 

\begin{figure*}[t!]
	\centering
	\includegraphics[width=0.96\textwidth]{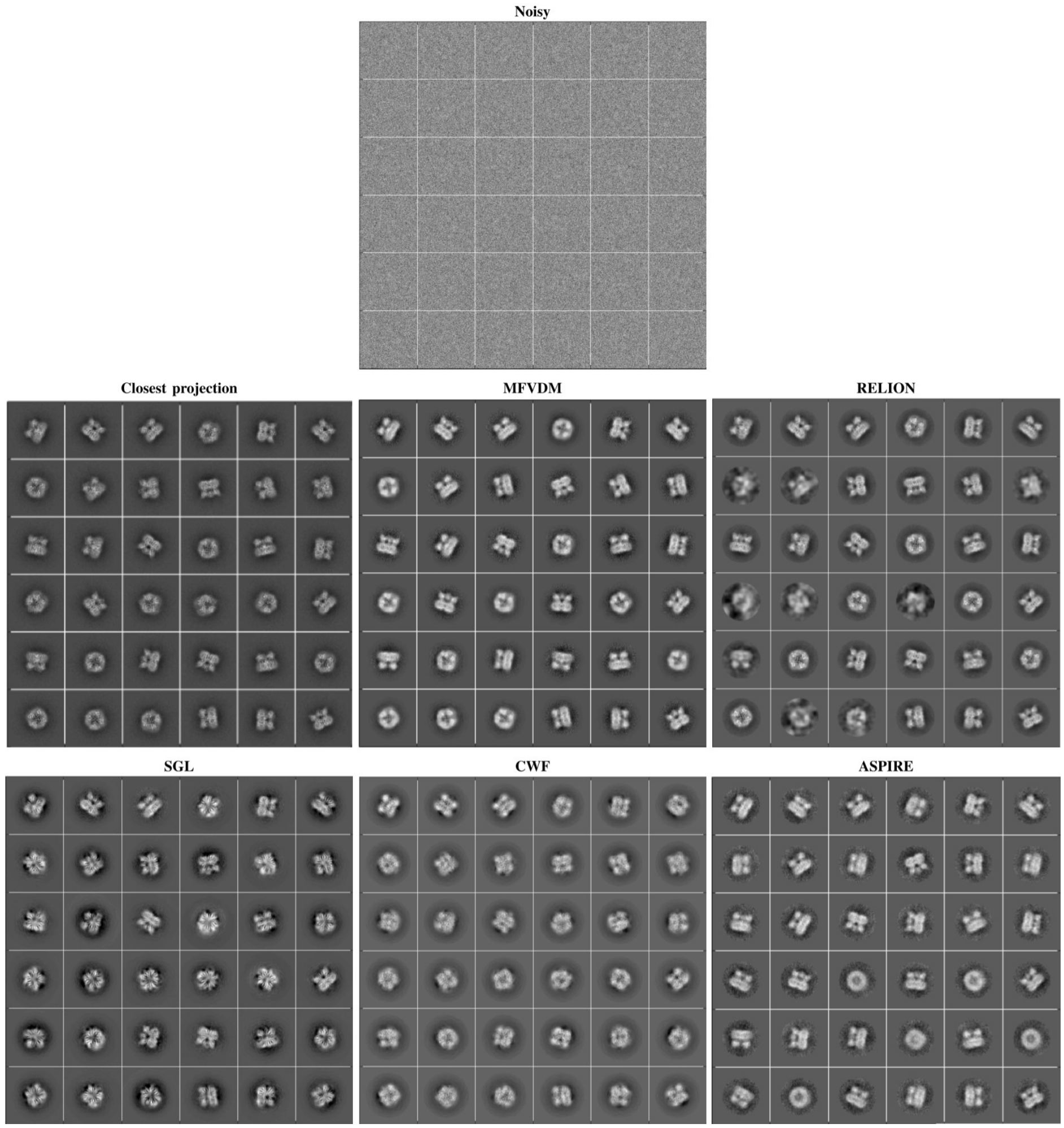}
    \caption{Image samples of TRPV1 dataset. \textbf{Noisy}: raw projection images.  \textbf{Closest projection: } the clean projection images from the reference map that have the highest correlation with the noisy images estimated by RELION. \textbf{MFVDM, RELION, SGL, CWF, ASPIRE: } the corresponding denoised images.}
    \label{fig:real_denoise_EM10005}
\end{figure*}

\begin{figure*}[t!]
	\centering
	\includegraphics[width=0.96\textwidth]{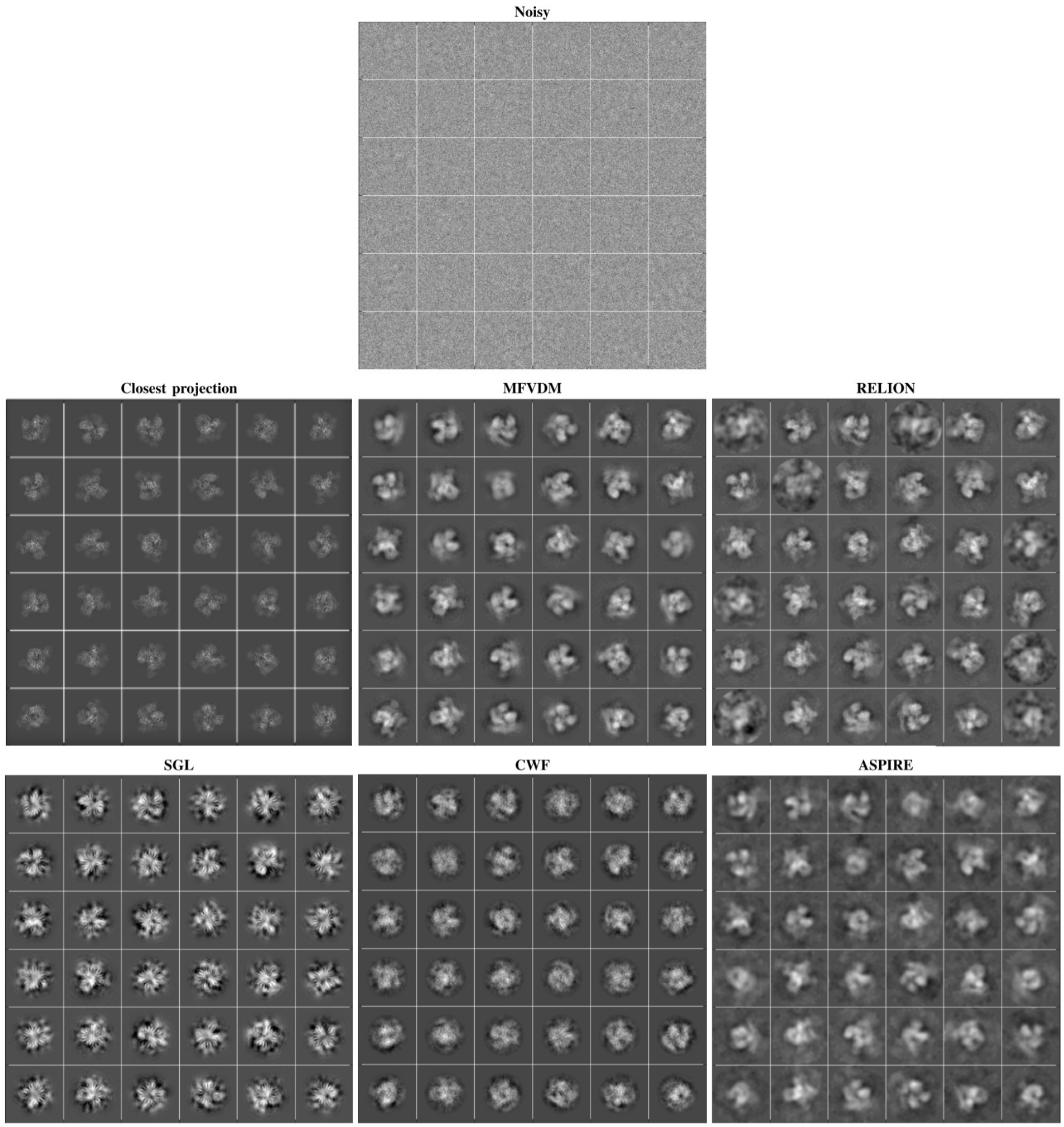}
    \caption{Image samples of HIV-1 Nef dataset. \textbf{Noisy}: raw projection images.  \textbf{Closest projection: } the clean projection images from the reference map that have the highest correlation with the noisy images estimated by RELION. \textbf{MFVDM, RELION, SGL, CWF, ASPIRE: } the corresponding denoised images. }
	\label{fig:real_denoise_EM10178}
\end{figure*}

\vspace{0.1cm}
\noindent\textbf{Experimental Datasets: }In Figs.~\ref{fig:real_denoise_EM10005} and~\ref{fig:real_denoise_EM10178} we show more denoising results for two experimental datasets: TRPV1 and HIV-1 Nef, by the following approaches: MFVDM, RELION 2D class averaging, SGL, CWF and ASPIRE 2D averaging. We generated the clean template images from the reference map and estimated the most similar template for each raw image. We used the 3D auto refine tool in RELION 2.1 to carry out the template matching (notice the closest projection may not be the truth beneath the raw image due to the high level of noise). The Figs.~\ref{fig:real_denoise_EM10005} and~\ref{fig:real_denoise_EM10178} show  that our proposed method (MFVDM) is able to recover all views. RELION is good at recovering a few views, however, some class-means have very poor quality. The MFVDM and RELION 2D class averaing outperform CWF, SGL and ASPIRE on these real datasets. %CWF and SGL perform not too bad on TRPV1 data, but have poor results on HIV1-Nef. For ASPIRE 2D class averaging, note that we apply a Gaussian filter with $\sigma = 2$ on ASPIRE denoised images. 
%For ASPIRE 2D class averaging, the main structures on both datasets are preserved, while some details are blurry or lost, e.g., the details on the right bottom two figures of TRPV1 data.

For the real image data, we cannot use the estimated closest projections to directly evaluate 2D image denoising in terms of MSE, SSIM and PSNR, because (1) the real single particle images are not perfectly centered, and (2) the estimated closest projections might be inaccurate. Therefore, the quantitative evaluation for real data is through the comparisons of the resolution of the reconstructed volume and the Fourier shell correlations shown in the main paper.

\end{document}